\documentclass{article}

\usepackage{CJKutf8} % 支持中文
\usepackage[utf8]{inputenc}

\usepackage[preprint,nonatbib]{neurips_2022}

\usepackage[utf8]{inputenc} % allow utf-8 input
\usepackage[T1]{fontenc}    % use 8-bit T1 fonts
\usepackage{hyperref}       % hyperlinks
\usepackage{url}            % simple URL typesetting
\usepackage{booktabs}       % professional-quality tables
\usepackage{amsfonts}       % blackboard math symbols
\usepackage{nicefrac}       % compact symbols for 1/2, etc.
\usepackage{microtype}      % microtypography
\usepackage{xcolor}         % colors

\usepackage{wrapfig}
\usepackage{times}
\usepackage{latexsym}

\definecolor{darkblue}{rgb}{0.0,0.0,0.5}

\usepackage{amsmath}
\usepackage{array}
\newcolumntype{L}{>{$}l<{$}}
\newcolumntype{C}{>{$}c<{$}}
\newcolumntype{R}{>{$}r<{$}}

\usepackage{multirow}
\usepackage{tabularx, caption, boldline}
\usepackage{graphicx}
\usepackage{cellspace}
\usepackage{algpseudocode}  
\usepackage{multicol}  
\usepackage{flexisym}

\usepackage{microtype}

\usepackage{dsfont}

\makeatletter
\def\hlinewd#1{%
\noalign{\ifnum0=`}\fi\hrule \@height #1 %
\futurelet\reserved@a\@xhline}
\makeatother

\usepackage{times}
\usepackage{latexsym}

\usepackage{algpseudocode}  
\usepackage{amsmath}
\usepackage{multicol}  
\usepackage{multirow} 
\usepackage{graphicx}  %Required
\usepackage{array}
\usepackage{arydshln}
\usepackage{booktabs}
\usepackage{textcomp}

\usepackage{amssymb}% http://ctan.org/pkg/amssymb
\usepackage{pifont}% http://ctan.org/pkg/pifont

\usepackage{cleveref}
\crefformat{section}{\S#2#1#3} % see manual of cleveref, section 8.2.1
\crefformat{subsection}{\S#2#1#3}
\crefformat{subsubsection}{\S#2#1#3}

\usepackage{microtype}

\usepackage{adjustbox}
\usepackage{multicol}  
\usepackage{multirow} 
\usepackage{amsmath}
\usepackage{amsfonts}
\usepackage{makecell}
\usepackage{algpseudocode} 
\usepackage{verbatim}
\usepackage{mathtools}
\DeclareMathOperator*{\argmax}{arg\,max}

\usepackage{booktabs}

\usepackage{makecell}
\usepackage{cleveref}
\usepackage[normalem]{ulem}

\usepackage{CJKutf8}

\usepackage{times}
\usepackage{latexsym}

\usepackage{footnote}
\makesavenoteenv{tabular}
\makesavenoteenv{table}

\usepackage[hang,flushmargin]{footmisc}

\usepackage[linesnumbered,ruled]{algorithm2e}
\SetAlFnt{\small}
\SetAlCapNameFnt{\normalsize}
\makeatletter
\newcommand{\nosemic}{\renewcommand{\@endalgocfline}{\relax}}% Drop semi-colon ;
\newcommand{\dosemic}{\renewcommand{\@endalgocfline}{\algocf@endline}}% Reinstate semi-colon ;
\let\oldnl\nl% Store \nl in \oldnl
\newcommand{\nonl}{\renewcommand{\nl}{\let\nl\oldnl}}% Remove line number for one line
\makeatother

\usepackage{enumitem}
\usepackage{tablefootnote}

\usepackage{tabularx}
\makeatletter
\def\hlinewd#1{%
\noalign{\ifnum0=`}\fi\hrule \@height #1 %
\futurelet\reserved@a\@xhline}
\makeatother

\usepackage{minitoc}
\usepackage{cleveref}
\crefformat{section}{\S#2#1#3}
\crefformat{subsection}{\S#2#1#3}
\crefformat{subsubsection}{\S#2#1#3}
\crefrangeformat{section}{\S\S#3#1#4 to~#5#2#6}
\crefmultiformat{section}{\S\S#2#1#3}{ and~#2#1#3}{, #2#1#3}{ and~#2#1#3}
\usepackage{refstyle}
\Crefformat{figure}{#2Figure~#1#3}
\Crefmultiformat{figure}{Figures.~#2#1#3}{ and~#2#1#3}{, #2#1#3}{ and~#2#1#3}
\Crefformat{table}{#2Table~#1#3}
\Crefmultiformat{table}{Tables~#2#1#3}{ and~#2#1#3}{, #2#1#3}{ and~#2#1#3}
\Crefformat{appendix}{Appx.~\S#2#1#3}
\crefformat{algorithm}{Alg.~#2#1#3}

\definecolor{NavyBlue}{rgb}{0.1, 0.4, 0.8}
\hypersetup{colorlinks = true, linkcolor = NavyBlue,
            urlcolor  = gray,
            citecolor = NavyBlue,
            anchorcolor = NavyBlue}

\title{Language Models Can See:\\ Plugging Visual Controls in Text Generation}

\author{
 \textbf{Yixuan Su}$^{\spadesuit,}$\thanks{Project Lead, \texttt{ys484@cam.ac.uk}}  \quad
 \textbf{Tian Lan}$^{\diamondsuit,}$\thanks{Co-second Authors}  \quad
 \textbf{Yahui Liu}$^{\clubsuit,}$\footnotemark[2]  \quad
 \textbf{Fangyu Liu}$^{\spadesuit,}$\footnotemark[2]  \quad
 \\
 \textbf{Dani Yogatama}$^\heartsuit$  \quad
 \textbf{Yan Wang}$^\diamondsuit$  \quad
 \textbf{Lingpeng Kong}$^\triangleright$ \quad
 \textbf{Nigel Collier}$^{\spadesuit,}$\thanks{Corresponding Author, \texttt{nhc30@cam.ac.uk}} \quad
 \\
 $^\spadesuit$University of Cambridge \ \ \ \ \ $^\diamondsuit$Tencent AI Lab \ \ \ \ \ $^\clubsuit$University of Trento\\
 $^\heartsuit$DeepMind \ \ \ \ $^\triangleright$The University of Hong Kong\\
}

\begin{document}

\maketitle

\doparttoc 
\faketableofcontents

\begin{abstract}
Generative language models (LMs) such as GPT-2/3 can be prompted to generate text with remarkable quality. While they are designed for text-prompted generation, it remains an open question how the generation process could be guided by modalities beyond text such as images. In this work, we propose a training-free framework, called MAGIC (i\textbf{\underline{MA}}ge-\textbf{\underline{G}}uided text generat\textbf{\underline{I}}on with \textbf{\underline{C}}LIP), for plugging in visual controls in the generation process and enabling LMs to perform multimodal tasks (e.g., image captioning) in a zero-shot manner. 
MAGIC is a simple yet efficient plug-and-play framework, which directly combines an off-the-shelf LM (i.e., GPT-2) and an image-text matching model (i.e., CLIP) for image-grounded text generation. During decoding, MAGIC influences the generation of the LM by introducing a CLIP-induced score, called \textit{magic score}, which regularizes the generated result to be semantically related to a given image while being coherent to the previously generated context. Notably, the proposed decoding scheme does not involve any gradient update operation, therefore being computationally efficient. On the challenging task of zero-shot image captioning, MAGIC outperforms the state-of-the-art method by notable margins with a nearly 27 times decoding speedup. MAGIC is a flexible framework and is theoretically compatible with any text generation tasks that incorporate image grounding. In the experiments, we showcase that it is also capable of performing visually grounded story generation given both an image and a text prompt. Our code, models and other related resources are publicly released at \url{https://github.com/yxuansu/MAGIC}.
\end{abstract}

%--------------------------%
\section{Introduction}
\label{sec:introduction}
%--------------------------%

Since the introduction of GPT-2 \cite{radford2019language}, generative language models (LMs), which are pre-trained on enormous amount of unstructured text, have produced unmatched performances on a wide range of NLP tasks \cite{brown2020language,chowdhery2022palm}. 
%After pretraining on enormous amount of unstructured text, in inference time, 
Given a textual prompt, LMs can continuously generate texts with the next-token prediction decoding scheme.  
%While controlling GPT models have become possible by inserting textual prompts,
Although controlling the outputs of LMs have become possible by inserting textual prompts, it is still unknown how the decoding process could be guided by information beyond texts, such as images. 

Recently, multimodal representation learning of text and images have been rejuvenated by %large-scale noisy image-text pairs 
pre-trained image-text joint embedding models, such as CLIP \cite{radford2021learning} and ALIGN \cite{jia2021scaling}. They leverage large-scale nosiy image-text pairs with weak correspondence for contrastive embedding learning and the learned joint model achieves impressive zero-shot performance competitive to supervised models on tasks such as image classification and image-text retrieval. However, they are still under-explored for image-grounded text generation.\footnote{Note that while such noisy weak image-text pair supervision is sufficient for learning embeddings, they could not be directly used to train image captioning model due to the data's extreme level of noise \cite{tewel2021zero}.} 

How can we combine the best of both the pre-trained LMs and image-text embedding models for visually grounded text generation? Existing supervised methods combine multimodal encoders by further training them on human-annotated paired image-text data \cite{Mokady2021ClipCapCP,chen2021visualgpt}.
Differently, weakly supervised approaches \cite{anderson2018partially,feng2019unsupervised,Laina2019TowardsUI} rely on pre-trained object detectors to identify visual concepts and create pseudo image-text pairs. Instead of training on annotated image-text pairs, they directly train on the pseudo data.  %\cite{anderson2018partially,feng2019unsupervised,Laina2019TowardsUI}. 
However, such methods are usually limited by the object detectors that are trained with a fixed set of labels.
The closest to our proposal is ZeroCap~\cite{tewel2021zero} which is an unsupervised image captioning method by combining frozen CLIP and GPT-2. One of the advantages of ZeroCap is it performs \emph{ex post facto} in the activation space without re-training or fine-tuning the CLIP and GPT-2 models. However, ZeroCap relies on gradient update and optimization over the context cache, which significantly slows down the inference and hindering its use in real-world scenarios.

In this paper, we propose to solve this challenging task in a completely new perspective 
%We approach image-guided text generation from a completely new angle 
by designing a novel text decoding scheme, called MAGIC (i\textbf{\underline{MA}}ge-\textbf{\underline{G}}uided text generat\textbf{\underline{I}}on with \textbf{\underline{C}}LIP). During inference, MAGIC does not rely on \textit{any} additional training or parameters and utilizes explicit ``control knobs'' to select desired outputs following the guidance of both the GPT-2 and CLIP models. 
%but directly combine a GPT-2 and a CLIP model for image-guided text generation. 
Different from the standard decoding process of GPT-2, we insert a CLIP-induced term, called \textit{magic score}, in the next token search to encourage the predicted result to demonstrate information that is close to a given image. Our experiments show that such a framework enables zero-shot image captioning and also visually grounded story generation under a simple plug-and-play principle. 

%We conduct comprehensive experiments to demonstrate the effectiveness and robustness of our proposed method. 
To verify the qualitative and quantitative performance of the proposed MAGIC method, we conduct comprehensive experiments on two commonly used benchmarks (Section \cref{sec:image_captioning}): MS-COCO~\cite{lin2014microsoft} and Flickr30k~\cite{plummer2015flickr30k}. To our surprise, MAGIC achieves state-of-the-art (SOTA) performance across different evaluation metrics, especially outperforming all unsupervised and weakly supervised baselines notably. Moreover, since MAGIC involves no gradient update, the inference speed accelerates upon previous zero-shot image captioning SOTA by around 27 times. Beyond image captioning, we also test our approach on visually grounded story generation (Section \cref{sec:story_generation}). In this task, given an image and a text prompt, MAGIC can generate high-quality stories that outperform strong baseline methods on both human and automatic evaluations.%human-evaluated metrics, including coherence and imaginability.

%In summary, we make the following contributions: 
%\begin{enumerate}
%    \item As far as we know, we are the first to propose a zero-shot method, %called MAGIC, to utilize explicit ``control knobs'' to efficiently sample %desired captions following the guidance of both the pre-trained GPT-2 and CLIP %models;
%    \item We empirically show that MAGIC is extremely effective on zero-shot %image captioning, achieving SOTA across different benchmarks;
%    \item We demonstrate that MAGIC could be used in creative ways: it can %perform complex multimodal generation tasks such as image-text prompted story %generation and reaches near-human performances on a wide range of metrics.
%\end{enumerate}

In summary, we make the following contributions: 
\begin{itemize}%[noitemsep,topsep=1pt]
    \item To the best of our knowledge, we are the first to propose a zero-shot method, called MAGIC, to utilize explicit ``control knobs'' to efficiently select desired outputs following the guidance of both the pre-trained GPT-2 and CLIP models;
    \item We empirically show that MAGIC is extremely effective on zero-shot image captioning, achieving SOTA across different benchmarks;
    \item We demonstrate that MAGIC could be used in creative ways: it can perform complex multimodal generation tasks such as visually grounded story generation and reaches near-human performances on a wide range of evaluation metrics.
\end{itemize}

%(1) we contribute a first-of-its-kind zero-shot method, called MAGIC, for controlling text generation with images; (2) while being extremely simple, we empirically show that MAGIC is extremely effective on zero-shot image captioning, achieving SOTA across different benchmarks; (3) we demonstrate that MAGIC could be used in creative ways: it can perform complex multimodal generation tasks such as text-image prompted story generation and reaches near-human performances on a wide range of metrics.

%The MAGIC framework opens a new avenue for cross-modal plug-and-play 

%We combine the best of both worlds, ..
%are great but they do not have access to modalities beyond text. 
%In parallel we have large-scale pretrained text-image embeddings but they can not do image captioning.
%How can we enable LMs to see without additional training / aligning embedding space? 
%We propose ...

%--------------------------%
\section{Background}
\label{sec:related-work}
%--------------------------%

In this section, we briefly introduce previous work related to our research. % both image captioning, and plug and play language models.
\subsection{Image Captioning}
Our work is closely related to the literature of image captioning, where the goal is to describe images with meaningful and syntactically correct sentences.
Although this topic has been extensively explored in the past few years, it is still far from being considered as a solved task. Given the training strategies (e.g., the type of training data), we can roughly classify the previous methods into two categories: (1) Supervised and (2) Weakly-/Un-Supervised approaches. The former heavily depends on manually labelled image-text datasets. 
In contrast, the latter tries to create pseudo image-text pairs 
%reduce the amount of paired text-image data 
(i.e., weakly supervised) or even avoid using any paired image-text data (i.e., unsupervised).

\noindent \textbf{Supervised Approaches.} With the development of deep learning, most of the existing models use one CNN to encode the input image and one RNN to generate the corresponding sentence describing the image~\cite{mao2014explain,vinyals2015show}. These models are trained to maximize the probability of generating the ground-truth captions conditioned on the input image. After that, the main focus of following methods is to model the interaction between visual and 
textual cues via attention mechanism to get more faithful and richer captions~\cite{xu2015show,lu2017knowing,anderson2018bottom,zhang2021rstnet,huang2021unifying}. Meanwhile, some controllable image captioning methods~\cite{mathews2016senticap,gan2017stylenet,chen2018factual,shuster2019engaging,chen2021human} propose to generate diverse descriptions by feeding different control signals (e.g., label and text), which require additional annotations for training. Beyond describing the whole image scene, dense captioning methods~\cite{johnson2016densecap,chatterjee2018diverse,kim2019dense,yin2019context,zeng2020dense} aim to describe the visual objects in a sub-region of the input image. Recently, vision-language pre-training methods~\cite{zhou2020unified,Li2020OscarOA,Mokady2021ClipCapCP,hu2021scaling}, benefiting from the rich visual-textual representation of pre-trained models on large-scale datasets, are tendencies for vision-language generation by re-training or fine-tuning the model parameters on downstream tasks. Although these methods have achieved impressive results, a certain amount of paired image-text data is indispensable during training. %Moreover, with limited annotated image-text data (e.g. limited amount of supervised “gold-labels” in MS-COCO~\cite{lin2014microsoft}), it is difficult to generalize to images in the wild, especially generating captions for novel objects that are not appeared in the datasets~\cite{mao2015learning,tran2016rich,hendricks2016deep,venugopalan2017captioning}.

\noindent \textbf{Weakly-/Un-Supervised Approaches.}
Till now, there has been several attempts to reduce the reliance on paired image-text data for the training of image captioning model. 
%For example,  \cite{zhao2017dual} use the cycle consistency objective~\cite{zhu2017unpaired} for the cross-domain problem.
In weakly-supervised approaches, employing \textit{pseudo-captions}, i.e., sentences that contain the object labels detected from the given images, has been a common choice \cite{anderson2018partially,feng2019unsupervised,Laina2019TowardsUI}.
%for training image captioning models without labeled image-text pairs~. 
%The main challenge here is to utilize 
However, a weakly supervised cross-modal alignment between image and text is needed. % leveraging visual concepts in the image. 
%On the one hand, 
Besides, \textit{pseudo-captions} tend to contain irrelevant words for the given images~\cite{Honda2021RemovingWS}. Therefore, it requires carefully designed constraints or learning schema to alleviate the noise. 
%On the other hand, 
What is more, such methods require a pre-trained object detector bounded by a fixed set of labels to provide visual concepts. They are thus ineffective for any out-of-domain concepts and scenes. 
%Intuitively, a visual encoder that is trained on more diverse and large-scale data sources, unbounded by a fixed set of labels, can be with better generalization ability to unseen objects and concepts.

How can we get rid of creating \textit{pseudo-captions} and perform image captioning in a truly unsupervised manner? Recently, CLIP~\cite{radford2021learning} has emerged as a successful vision-language alignment model by training on 400M noisy web-collected image-sentence pairs. It has shown impressive zero-shot capabilities on various vision-language tasks and can open new avenues for answering the former question. ZeroCap~\cite{tewel2021zero} is the most related to our work. It is built on a pre-trained CLIP model together with the GPT-2 language model~\cite{radford2019language}. Different from previous work, ZeroCap is truly zero-shot, where the optimization is performed ``\textit{ex post facto}'' in the activation space without re-training or fine-tuning the model parameters. In ZeroCap, the whole context cache (i.e., all the $K$ and $V$ in the self-attention modules~\cite{vaswani2017attention,dosovitskiy2021image}) is updated with the guidance of CLIP and GPT-2 for every prediction step. As a result, the computational overhead of such optimization steps will increase drastically as the size of language model goes up. One key difference of our proposal  with respect to ZeroCap is that MAGIC 
%we propose a novel and efficient decoding method, named as MAGIC Search (See Section~\ref{sec:magic-search}), which 
utilizes explicit ``control knobs'' to select desired outputs corresponding to the given image. Since our procedure does not involve any gradient updating or optimization, it significantly speeds up the decoding process by around 27 times (Section~\cref{sec:image_caption_result}). 

%\yahui{each captioning dataset incorporates a different style, which may not be natural for the pre-trained language model~\cite{Mokady2021ClipCapCP}. The motivation for fine-tuning GPT-2 on the captioning text only.}

%//TODO highlight the differences between ZeroCap and Ours method.

%\subsection{Vision-Language Pretraining}

\subsection{Plug and Play Generative Models}
\label{sec:plug_and_play_related_work}
Lagre-scale pre-trained generative models have revolutionized the field of natural language processing~\cite{radford2019language,brown2020language} and computer vision~\cite{radford2021learning,ramesh2021zero,ramesh2022hierarchical,karras2019style,karras2020analyzing,karras2021alias} in the past few years. Various previous work~\cite{nguyen2016synthesizing,nguyen2017plug,dathathri2020plug,shen2020interpreting} have revealed that there are rich meaningful semantics in the features learned by such models. This shows a promising pathway to synthesize the desired outputs by interpreting the existing generative models in a ``\textit{plug and play}'' manner. 
%Obviously, one of he greatest strengths is that, 
We can then directly enjoy the powerful capabilities of these \textit{off-the-shelf} big models (without any re-training or fine-tuning) and focus on the lightweight task-specific optimization.

For instance, in the image generation field, DGN-AM~\cite{nguyen2016synthesizing} can generate images conditioned on a class by finding a hidden code %that the generator maps to an image 
that clearly activates a neuron in another classifier. Then, PPGN~\cite{nguyen2017plug} improves the diversity and quality of the synthesized images by incorporating approximate Metropolis-adjusted Langevin (MALA) algorithm~\cite{roberts1996exponential,roberts1998optimal}. Shen \emph{et al.}~\cite{shen2020interpreting} propose to directly travel in the latent space of pre-trained unconditional GANs to manipulate the attributes of the input image. Patashnik \emph{et al.}~\cite{patashnik2021styleclip} use CLIP to connect the text prompt and images to search the latent codes of StyleGAN by gradient descent optimization, which finally results in the manipulation of the visual attributes in the input image. Similarly, in the text generation field, PPLM~\cite{dathathri2020plug} extends the previous PPGN to text generation tasks (i.e., editing topic and sentiment), where the image generative models is replaced with a GPT-2 language model. 
Most recently, ZeroCap~\cite{tewel2021zero} shows image captioning task can be tackled by directly combining the existing CLIP and GPT-2 models. In general, most of these mentioned ``plug and play'' methods 
%(except ~\cite{shen2020interpreting})
require iteratively shifting the hidden code (or latent code, or context cache) with gradient descent optimization. 

Different from previous work, our proposed approach extends the ``plug and play'' paradigm by optimizing the decoding strategy of generative LMs. Since MAGIC does not involve any gradient update in the inference, it is computationally efficient (e.g., $\sim$27 times faster than ZeroCap). Notably, although GPT-2 could generate synthetic text samples of unprecedented quality, it may not be natural on some task-specific text~\cite{Mokady2021ClipCapCP,shen2021much,zhang2021vinvl}. Following this observation, we continue fine-tuning the GPT-2 model on the task-specific text corpus in an unsupervised manner to adapt it to the textual domain of the end task~\cite{Laina2019TowardsUI,Honda2021RemovingWS}. The computational consumption of such adaptation is negligible (e.g., less than 2 hours with 1 NVIDIA 1080Ti GPU on MS-COCO). 
%to adapt the captioning text domain (See Section~\cref{sec:language_modelling}), 
During decoding, the fine-tuned GPT-2 and CLIP models are fixed. 
%MAGIC is well aligned with the ``plug and play'' paradigm.

%\yahui{Motivation: To utilize the visual controls in text generation, we need a transformer-based LM (e.g., GPT-2), which models the probability inference for the $i$-th word $x_i$ of the sentence $\pmb{x}$, i.e., $p(x_i \| \mathbf{x}_{<i})$.  As well known, GPT-2 is a language model trained on extremely large amounts of Internet text. Although it shows the ability to generate conditional synthetic text samples of unprecedented quality, it may be be not natural on some task-specific text~\cite{Mokady2021ClipCapCP}. Hence,  we introduce a low-energy fine-tuning procedure that involves two objectives: (1) adapting to the task-specific text domain, and (2) generating more informative text. Then, we introduce the two losses $\mathcal{L}_{\text{MLE}}$ and $\mathcal{L}_{\text{CL}}$ ...}

%--------------------------%
\section{Methodology}
\label{sec:methodology}
%--------------------------%

%--------------------------%
\subsection{Unsupervised Language Modelling}
\label{sec:language_modelling}
%--------------------------%

Following previous studies \cite{Laina2019TowardsUI,Honda2021RemovingWS}, we first learn an unsupervised language model on the text corpus of the end task 
%of interest to create a meaningful 
to adapt to its textual domain. 
%The goal of language modelling is to learn a probability distribution $p_{\theta}$ over an unstructured text corpora $\mathcal{D}=\{\boldsymbol{x}_n\}_{n=1}^{|\mathcal{D}|}$, where $\boldsymbol{x}_i$ is a variable-length text sequence and $\theta$ denotes the parameters of the language model.
Typically, given a variable-length text sequence $\boldsymbol{x}$, the maximum likelihood estimation (MLE) objective is used to train the language model $\theta$ as 
\begin{equation}
    \label{eq:mle}
    \mathcal{L}_{\textup{MLE}} = -\frac{1}{|\boldsymbol{x}|}\sum_{i=1}^{|\boldsymbol{x}|}\log p_{\theta}(x_i|\boldsymbol{x}_{<i}).
\end{equation}
Recently, Su \emph{et al.}~\cite{DBLP:journals/corr/abs-2202-06417} propose to incorporate contrastive objective into the training of the language model to calibrate the  model's representation space and obtain better language model perplexity. Given the text sequence $\boldsymbol{x}$, the contratsive objective $\mathcal{L}_{\textup{CL}}$ is defined as  
\begin{equation}
    \label{eq:cl}
    \mathcal{L}_{\textup{CL}} = \frac{1}{|\boldsymbol{x}|\times(|\boldsymbol{x}| - 1)}\sum_{i=1}^{|\boldsymbol{x}|}\sum_{j=1,j\neq i}^{|\boldsymbol{x}|}\max\{0,\rho - s(h_{x_i}, h_{x_i}) + s(h_{x_i}, h_{x_j})\},
\end{equation}
where $\rho$ is a pre-defined margin that regularizes the distribution of the model's representation space. The $h_{x_i}$ is the representation of token $x_i$ and
the similarity function $s$ computes the cosine similarity between token representations as $s(h_{x_i}, h_{x_j}) = h_{x_i}^\top h_{x_j}/(\|h_{x_i}\|\cdot\|h_{x_j}\|)$.
%\begin{equation}
%    \label{eq:cosine}
%    s(h_{x_i}, h_{x_j}) = \frac{h_{x_i}^\top h_{x_j}}{\|h_{x_i}\|\cdot\|h_{x_j}\|}.
%\end{equation}
%\fangyu{shall we unify the notation here. e.g. all vectors use \texttt{boldsymbol}?}

The overall learning objective $\mathcal{L}$ of the language model is then defined as 
\begin{equation}
    \label{eq:simctg}
    \mathcal{L} = \mathcal{L}_{\textup{MLE}} + \mathcal{L}_{\textup{CL}}.
\end{equation}

%for different downstream tasks, we 
%--------------------------%
\subsection{MAGIC Search}
\label{sec:magic-search}
%--------------------------%

We propose a new decoding scheme, \textit{MAGIC Search}, which aims to steer the decoding process of the language model towards a desired visual direction. Formally, given a text prefix $\boldsymbol{x}_{<t}$ and an image $\mathcal{I}$, the selection of the output token $x_t$ at time step $t$ follows
\begin{align}
\begin{split}
    \label{eq:score}
    x_t = \argmax_{v\in V^{(k)}}\bigg\{(1 - \alpha) \times &\underbrace{p_{\theta}(v|\boldsymbol{x}_{<t})}_{\textup{model confidence}} - \\
    \alpha \times \underbrace{(\max\{s(h_v, h_{x_j}):1\leq j \leq t-1\})}_{\textup{degeneration penalty}} &\: + \: \beta \times \underbrace{f(v|\mathcal{I}, \boldsymbol{x}_{<t}, V^{(k)})}_{\textup{magic score}} \bigg\},\\
\end{split}
\end{align}
where $V^{(k)}$ is the set of top-$k$ predictions from the model's probability distribution $p_{\theta}(\cdot|\boldsymbol{x}_{<t})$ 
%is the set of top-$k$ prediction
%$k$ most-probable candidates predicted by the model 
and $s$ is described in Section \cref{sec:language_modelling}. $h_v$ is the representation of the candidate token $v$ which is computed by the model given the concatenation of $\boldsymbol{x}_{<t}$ and $v$. Inspired by Su \emph{et al.}~\cite{DBLP:journals/corr/abs-2202-06417}, we incorporate the model confidence and degeneration penalty into Eq.~(\ref{eq:score}) to let the model decode the likely output while avoiding the model degeneration problem. %~\cite{DBLP:journals/corr/abs-2202-06417}. 
%We refer the readers to Su \emph{et al.}~\cite{DBLP:journals/corr/abs-2202-06417} for more details.

Meanwhile, we introduce a novel scoring criterion, \textit{magic score}, to plug in visual controls into the decoding process. Given the candidate $v$, the prefix $\boldsymbol{x}_{<t}$, and the image $\mathcal{I}$, the magic score is defined as the distribution of image-text similarity over the candidate set $V^{(k)}$. We build our image-text similarity measurement with a pre-trained CLIP model and the magic score is then defined as
\begin{equation}
    f(v|\mathcal{I}, \boldsymbol{x}_{<t}, V^{(k)}) = \frac{e^{\textup{CLIP}(\mathcal{I},[\boldsymbol{x}_{<t}:v])}}{\sum_{z\in V^{(k)}}e^{\textup{CLIP}(\mathcal{I},[\boldsymbol{x}_{<t}:z])}}=\frac{e^{h_{\mathcal{I}}^\top h_{[\boldsymbol{x}_{<t}:v]}}}{\sum_{z\in V^{(k)}}e^{h_{\mathcal{I}}^\top h_{[\boldsymbol{x}_{<t}:z]}}},
\end{equation}
where $h_{\mathcal{I}}$ is the image embedding of $\mathcal{I}$ produced by the CLIP image encoder. The $h_{[\boldsymbol{x}_{<t}:v]}$ is the text embedding of the sequence $[\boldsymbol{x}_{<t}:v]$ produced by the CLIP text encoder and $[:]$ denotes the concatenation operation. Intuitively, the magic score encourages the language model to generate text that is semantically related to the image content and the strength of the visual control is regulated by the hyper-parameter $\beta$ in Eq.~(\ref{eq:score}). When $\beta=0$, the visual control is disabled and MAGIC Search degenerates to the vanilla contrastive search \cite{DBLP:journals/corr/abs-2202-06417}. 

It should be emphasized that MAGIC Search allows us to directly plug visual controls into the decoding process of the language model, without the need of extra supervised training \cite{dathathri2020plug} or gradient update on additional features \cite{dathathri2020plug,tewel2021zero}. This property makes our method much more computationally efficient than previous approaches as demonstrated in our experiments (Section~\cref{sec:image_caption_result}).

%--------------------------%
\section{Zero-Shot Image Captioning}
\label{sec:image_captioning}
%--------------------------%
We first evaluate our approach on the task of zero-shot image captioning. 

\textbf{Evaluation Benchmarks.} We conduct experiments on two widely used benchmarks: MS-COCO \cite{lin2014microsoft} and Flickr30k \cite{plummer2015flickr30k}. For both datasets, we set up the training, validation, and test splits according to the protocols provided by Karpathy \emph{et al.}~\cite{karpathy2015deep}. 

\textbf{Implementation Details.} As described in Section~\cref{sec:language_modelling}, for each benchmark, we fine-tune the GPT-2 model on the training text corpus for 3 epochs and the contrastive loss margin $\rho$ in Eq. (\ref{eq:simctg}) is set as 0.5. We optimize the model with the Adam optimizer \cite{DBLP:journals/corr/KingmaB14} and a learning rate of 2e-5. Notably, this fine-tuning procedure is computationally negligible, i.e., less than 2 hours with 1 NVIDIA 1080Ti GPU. During decoding, the generation of the language model starts with a special start-of-sequence (i.e., \texttt{[sos]}) token. For MS-COCO, we set the $k$, $\alpha$, and $\beta$ in MAGIC Search (i.e., Eq.~(\ref{eq:score})) as 45, 0.1, and 2.0 based on the model's performance on the validation set. As for Flickr30k, these values are set as 25, 0.1, and 2.0, respectively.\footnote{In Appendix \ref{appendix:ablation_study}, we provide detailed ablation studies on the effect of different hyper-parameter setups.}

\textbf{Baselines.} We include several zero-shot methods as our baselines. (1) We compare the generated results of the language model by starting from the start-of-sequence (i.e., \texttt{[sos]}) token with different decoding methods, including top-$k$ sampling \cite{DBLP:conf/acl/LewisDF18} with $k=40$ and nucleus sampling \cite{DBLP:conf/iclr/HoltzmanBDFC20} with $p=0.95$. Moreover, we include contrastive search \cite{DBLP:journals/corr/abs-2202-06417} using the same $k$ and $\alpha$ as in MAGIC Search to see the direct effect of the proposed magic score (Eq.~(\ref{eq:score})).\footnote{Note that, our proposed MAGIC Search is equivalent to the contrastive search when $\beta$ in Eq.~(\ref{eq:score}) equals to 0.} Note that, these methods do \textbf{not} take into account the image input, therefore can be used to assess the performance lower-bound of the language model. (2) We also compare with a CLIP-based method, called CLIPRe. Given an image, it retrieves the most related caption from the training text corpus based on the image-text similarity as measured by CLIP. (3) Lastly, we compare with the current state-of-the-art approach, ZeroCap~\cite{tewel2021zero}, which performs CLIP-guided gradient update on the language model features during the decoding process. For a fair comparison, we use the same language model for ZeroCap as in our approach.

\textbf{Evaluation Metrics.} Following the common practice in the literature, we perform evaluation using BLEU-1 (B@1), BLEU-4 (B@4) \cite{papineni2002bleu}, METER (M) \cite{denkowski2014meteor}, ROUGE-L (R-L) \cite{lin2004automatic}, CIDEr \cite{vedantam2015cider}, and SPICE \cite{anderson2016spice}. In addition, we compare the relative decoding speed of our approach against other generation-based baselines. Here, the decoding speed is measured from the average inference time per image instance.\footnote{The decoding speed of different methods are measured on the same hardware platform with a batch size of 1.}

\subsection{Results}
\label{sec:image_caption_result}
Table \ref{tb:main_result} shows the results on zero-shot image captioning. For a comprehensive comparison, we also include the results of several representative (1) supervised methods: BUTD \cite{anderson2018bottom}, GVD \cite{zhou2019grounded}, UniVLP \cite{zhou2020unified}, ClipCap \cite{Mokady2021ClipCapCP}, Oscar \cite{Li2020OscarOA}, and LEMON \cite{hu2021scaling}; and (2) weakly supervised methods: UIC \cite{feng2019unsupervised}, IC-SME \cite{Laina2019TowardsUI}, S2S-SS and S2S-GCC \cite{Honda2021RemovingWS}.

From the results of Top-$k$, Nucleus, and Contrastive, we see that solely using the unsupervised language model without conditioning on image inputs can hardly generate meaningful captions.\footnote{For stochastic sampling methods (i.e., Top-$k$ and Nucleus), we report the results averaged over 3 runs with different random seeds. We refer to Appendix \ref{appendix:zero_shot_image_captioning} for more details on the numerical results.} On the other hand, the results of CLIPRe show that the ability of measuring image-text similarity enables CLIP to retrieve captions that better correlate with the test image from the training text corpus. However, the performance of CLIPRe still lags behind the current SOTA method, ZeroCap, by a large margin due to the data discrepancy between the training and test sets. Lastly, we observe that, on both benchmarks, MAGIC achieves the best performance on 11 out of 13 metrics, demonstrating the clear advantages of our proposed approach. Notably, while outperforming ZeroCap on 12 out of 13 metrics, MAGIC achieves a nearly 27$\times$ decoding speedup. This is because, during the decoding process, MAGIC does not involve any computationally inefficient operations like gradient updates \cite{dathathri2020plug,tewel2021zero}, which further validates the practical usage of our approach.

\begin{table*}[t]
    \small
	\centering  % 表居中
	\renewcommand{\arraystretch}{1.2}
	\setlength{\tabcolsep}{6pt}
	\scalebox{0.85}{
	\begin{tabular}{cccccccccccccc}
		\hlinewd{0.75pt}
		\multirow{2}{*}{\textbf{Model}}&\multicolumn{6}{c}{MS-COCO}&\multicolumn{6}{c}{Flickr30k}&\multirow{2}{*}{Speed}\\
		\cmidrule(lr){2-7}
		\cmidrule(lr){8-13}
		&B@1&B@4&M&R-L&CIDEr&SPICE&B@1&B@4&M&R-L&CIDEr&SPICE&\\
		\hline
		&\multicolumn{12}{c}{\textit{Supervised Approach}}&\\
		\hline
		%&&&&&&&&&&&&\\
		BUTD&77.2&36.2&27.0&56.4&113.5&20.3&-&27.3&21.7&-&56.6&16.0&-\\
		GVD&-&-&-&-&-&-&66.9&27.3&22.5&-&62.3&16.5&-\\
		UniVLP&-&36.5&28.4&-&116.9&21.2&-&30.1&23.0&-&67.4&17.0&-\\
		ClipCap&-&33.5&27.5&-&113.1&21.1&-&-&-&-&-&-&-\\
		Oscar&-&36.5&30.3&-&123.7&23.1&-&-&-&-&-&-&-\\
		LEMON&-&40.3&30.2&-&133.3&23.3&-&-&-&-&-&-&-\\
		%OFA&-&43.5&31.9&-&149.6&26.1&&&&&\\
		\hline
		&\multicolumn{12}{c}{\textit{Weakly Supervised Approach}}&\\
		\hline
		UIC&41.0&5.6&12.4&28.7&28.6&8.1&-&-&-&-&-&-&-\\
		IC-SME&-&6.5&12.9&35.1&22.7&-&-&7.9&13.0&32.8&9.9&-&-\\
		S2S-SS&49.5&6.3&14.0&34.5&31.9&8.6&-&-&-&-&-&-&-\\
		S2S-GCC&50.4&7.6&13.5&37.3&31.8&8.4&-&-&-&-&-&-&-\\
		\hline
		&\multicolumn{12}{c}{\textit{Unsupervised Approach}}&\\
		\hline
		Top-$k$&33.6&2.4&8.3&25.6&3.8&1.7&34.0&2.9&9.0&24.4&3.3&2.7&69.9$\times$\\
		Nucleus&32.6&2.3&7.8&24.8&3.1&1.4&32.6&2.4&8.1&23.4&2.5&2.4&\textbf{72.5}$\times$\\
		Contrastive&39.5&3.0&10.8&30.8&7.7&2.9&37.6&4.3&9.8&25.7&8.9&4.6&50.4$\times$\\
		CLIPRe&39.5&4.9&11.4&29.0&13.6&5.3&38.5&5.2&11.6&27.6&10.0&5.7&-\\
		ZeroCap&49.8&7.0&15.4&31.8&34.5&9.2&\textbf{44.7}&5.4&11.8&27.3&16.8&6.2&1.0$\times$\\
		\hline
		MAGIC&\textbf{56.8}&\textbf{12.9}&\textbf{17.4}&\textbf{39.9}&\textbf{49.3}&\textbf{11.3}&44.5&\textbf{6.4}&\textbf{13.1}&\textbf{31.6}&\textbf{20.4}&\textbf{7.1}&26.6$\times$\\
		\hlinewd{0.75pt}
	\end{tabular}}
    \caption{Image Captioning Results on MS-COCO and Flickr30k.}
    	\vspace{-1.5mm}
	\label{tb:main_result}
\end{table*}

\begin{table*}[h]
    \small
	\centering  % 表居中
	\renewcommand{\arraystretch}{1.2}
	\setlength{\tabcolsep}{6pt}
	\scalebox{0.9}{
	\begin{tabular}{ccccccccccccc}
		\hlinewd{0.75pt}
		\multirow{2}{*}{\textbf{Model}}&\multicolumn{6}{c}{MS-COCO $\Longrightarrow$ Flickr30k}&\multicolumn{6}{c}{Flickr30k $\Longrightarrow$ MS-COCO}\\
		\cmidrule(lr){2-7}
		\cmidrule(lr){8-13}
		&B@1&B@4&M&R-L&CIDEr&SPICE&B@1&B@4&M&R-L&CIDEr&SPICE\\
		\hline
		Top-$k$&34.9&2.4&7.5&24.2&2.3&1.7&30.0&1.8&8.5&23.6&2.5&1.7\\
		Nucleus&33.4&1.7&7.0&23.3&1.8&1.3&29.1&1.6&8.0&22.9&2.1&1.6\\
		Contrastive&40.3&5.3&10.7&30.5&5.1&3.4&33.8&3.2&10.2&25.5&4.2&3.7\\
		CLIPRe&38.7&4.4&9.6&27.2&5.9&4.2&31.1&3.0&9.9&22.8&8.5&3.9\\
		\hline
		MAGIC&\textbf{46.4}&\textbf{6.2}&\textbf{12.2}&\textbf{31.3}&\textbf{17.5}&\textbf{5.9}&\textbf{41.4}&\textbf{5.2}&\textbf{12.5}&\textbf{30.7}&\textbf{18.3}&\textbf{5.7}\\
		\hlinewd{0.75pt}
	\end{tabular}}
    \caption{Cross-Domain Evaluation. X $\Longrightarrow$ Y  means source domain $\Longrightarrow$ target domain.}
    	\vspace{-1.5mm}
	\label{tb:cross_domain_result}
\end{table*}

\subsection{Cross-Domain Experiment}
\label{sec:cross_domain_image_captioning}
To test the generalization ability of our approach, we conduct a cross-domain experiment. Specifically, we apply the unsupervised language model fine-tuned on the training text corpus of the source domain (e.g., MS-COCO) to perform inference on the test set of the target domain (e.g., Flickr30k). We compare MAGIC with several zero-shot methods, including Top-$k$, Nucleus, Contrastive, and CLIPRe.\footnote{Due to its extremely high computational overhead, we do not include ZeroCap in this experiment.} For CLIPRe, given a test image from the target domain, it retrieves the most related caption from the training text corpus of the source domain.

Table \ref{tb:cross_domain_result} shows the results on cross-domain evaluation, where we observe performance drops in all methods as compared with the in-domain evaluation results shown in Table \ref{tb:main_result}.\footnote{For stochastic sampling methods (i.e., Top-$k$ and Nucleus), we report the results averaged over 3 runs with different random seeds. We refer to Appendix \ref{appendix:zero_shot_image_captioning} for more detailed numerical results.} Nonetheless, MAGIC still performs the best among all compared methods, demonstrating its clear advantages in terms of robustness and generalization ability.
%\fangyu{The cross-domain experiments is using data from the same task but different datasets. It might be good to explain this more clearly and acknowledge that without finetuning on task-specific texts, the model can fail (can have some numbers show that finetune on e.g. wiki text but test on COCO captioning). That said, as long as it is text from the same task it still can work. This could be in appendix and we cite appendix here.}
\begin{figure*}[t] 
  \centering    
  \setlength{\abovecaptionskip}{3pt}
  \includegraphics[width=0.99\textwidth]{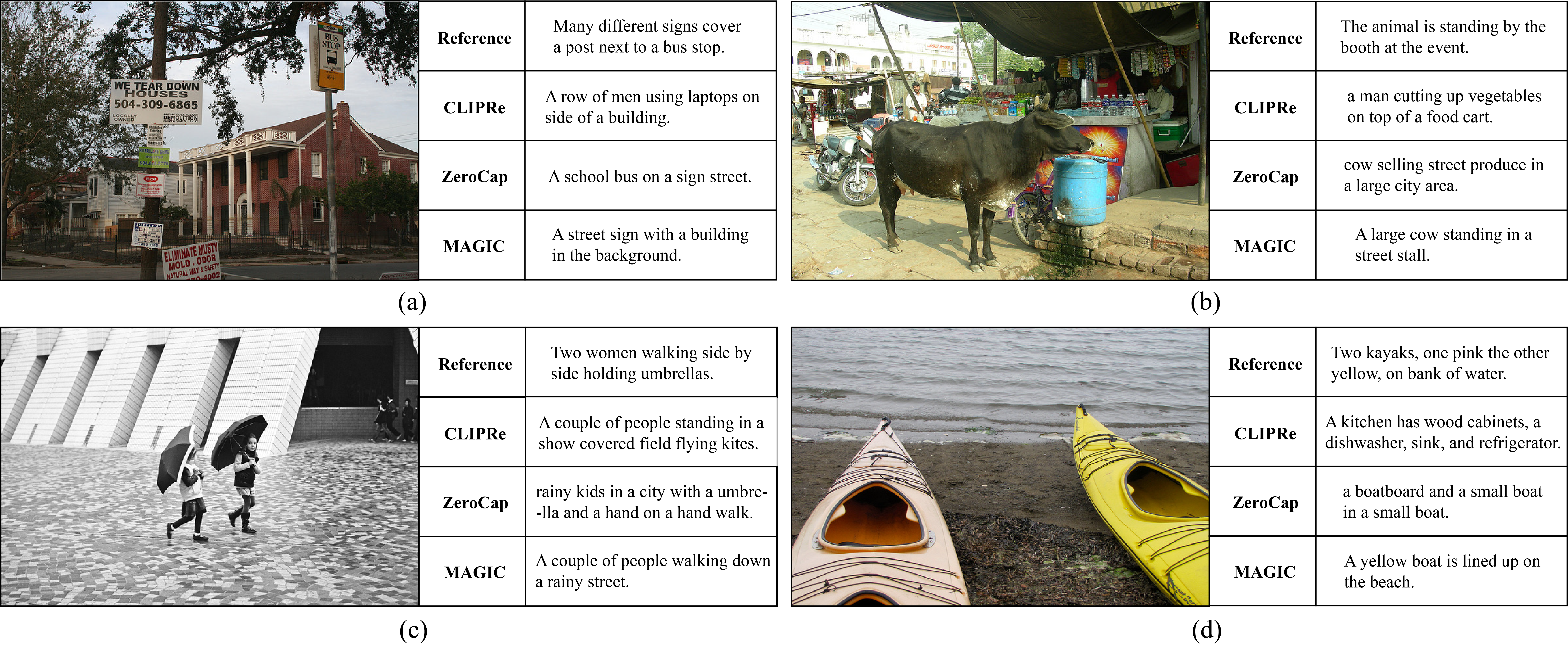}
  \caption{Examples of zero-shot image captioning. (Best viewed by zooming in.)}
  \label{fig:image_caption_case_study}
  \vspace{-1.5mm}
\end{figure*}

\subsection{Qualitative Evaluation}
Figure \ref{fig:image_caption_case_study} shows visual comparisons between our approach and other two strong zero-shot baselines along with the reference caption.\footnote{More examples of zero-shot image captioning are provided in Appendix \ref{appendix:image_captioning}.} The results demonstrate that MAGIC can generate fluent captions while being more effective at grounding on the given image. For example, in Figure \ref{fig:image_caption_case_study}(a), the result of CLIPRe only contains the object ``building'' that is partially related to the image. As for ZeroCap, it erroneously elaborates the object ``school bus'' which is not displayed in the image. On the other hand, MAGIC correctly describes the ``street sign'' object in the image as well as the building in the background. Next, we take Figure \ref{fig:image_caption_case_study}(d) as an example, in which the result of CLIPRe is clearly irrelevant to the image. As for ZeroCap, while it produces objects like ``boatboard'' and ``small boat'' that are related to the image, the generated result is not fluent and ungrammatical. In contrast, MAGIC is able to describe the correct objects such as ``yellow boat'' and ``beach'' as well as their positional relationship (i.e., lined up on) while maintaining the fluency and grammaticality of the generated text.

%--------------------------%
\section{Story Generation}
\label{sec:story_generation}
%--------------------------%
To verify the versatility and extensibility of MAGIC, we test it on a popular text generation task, i.e., \textit{story generation}. In this task, given a story title (i.e., text prompt), the language model is asked to generate an interesting and coherent story that is related to the story title. 
%coherent and interesting story.
%To verify the versatility and extensibility of our approach, we conduct experiments on another popular text generation task, i.e., \textit{story generation}. In this task, given a story title (i.e., text prompt), the model is asked to generate a coherent and interesting story.

\textbf{Evaluation Benchmark.} We evaluate our approach on the widely used ROCStories \cite{mostafazadeh2016corpus} dataset. In this dataset, each story title is accompanied with a five-sentence commonsense story written by human. We create the training, validation, and test sets following the official split.

\textbf{Model and Baselines.} We use a publicly available GPT-based language model \cite{DBLP:journals/corr/abs-2202-06417} which is fine-tuned on the ROCStories benchmark.\footnote{\url{https://huggingface.co/cambridgeltl/simctg_rocstories}} As MAGIC Search is a language model decoding scheme, we compare it with a range of strong text decoding methods, including (1) Greedy search; (2) Beam search with beam width of $10$; (3) Top-$k$ sampling \cite{DBLP:conf/acl/LewisDF18} with $k=40$; (4) Nucleus sampling \cite{DBLP:conf/iclr/HoltzmanBDFC20} with $p=0.95$; (5) Typical sampling \cite{meister2022typical} with $\tau=0.2$; and (6) Contrastive search \cite{DBLP:journals/corr/abs-2202-06417} with $k=5$ and $\alpha=0.6$. The hyperparameters of different methods are selected based on their optimal MAUVE \cite{pillutla2021mauve} (detailed in Section \cref{sec:story_generation_automatic_evaluation}) performance on the validation set.

%All baseline hyperparameters are selected based on their performance on the validation set.

\textbf{Implementation Details of MAGIC.} To perform MAGIC Search, given the story title, we first retrieve the image (from an image index) that is most related to the story title as measured by CLIP. We construct the image index with the public ConceptualCaptions \cite{sharma2018conceptual} dataset that contains over 3.3M images from the web. In practice, we pre-compute the image representations with CLIP and build the image index with FAISS \cite{johnson2019billion}, therefore supporting a fast ``story title-image'' retrieval with sub-linear time complexity. Then, by visually grounding on the retrieved image, we generate the story from the story title using MAGIC Search ($k=5$, $\alpha=0.6$, and $\beta=0.15$).\footnote{The hyper-parameters are
selected based on the model's optimal MAUVE performance on the validation set.}

\begin{table*}[t]
    \small
	\centering  % 表居中
	\renewcommand{\arraystretch}{1.2}
	\setlength{\tabcolsep}{6pt}
	\scalebox{0.84}{
	\begin{tabular}{cccccccccccc}
		\hlinewd{0.75pt}
		\multirow{2}{*}{\textbf{Method}}&\multicolumn{7}{c}{Automatic Evaluation}&\multicolumn{4}{c}{Human Evaluation}\\
		\cmidrule(lr){2-8}
		\cmidrule(lr){9-12}
        &rep-2$\downarrow$&rep-3$\downarrow$&rep-4$\downarrow$&div.$\uparrow$&coh.$\uparrow$&MAUVE$\uparrow$&CLIPScore$\uparrow$&coh.$\uparrow$&flu.$\uparrow$&inform.$\uparrow$&si-rel.$\uparrow$\\
        \hlinewd{0.75pt}
        Agreement&-&-&-&-&-&-&-&0.68&0.57&0.66&0.73\\
        \hline
        Greedy&22.27&15.42&12.36&0.58&0.473&0.53&0.23&2.67&3.20&3.10&2.03\\
        Beam&26.76&21.79&18.85&0.47&0.478&0.46&0.25&2.71&3.23&3.15&2.05\\
        Top-$k$&3.38&0.76&0.23&0.95&0.458&0.86&0.21&2.52&3.69&3.62&1.96\\
        Nucleus&2.92&0.60&0.18&0.96&0.452&0.88&0.21&2.48&3.68&3.71&1.92\\
        Typical&2.52&0.46&0.12&\textbf{0.97}&0.450&0.84&0.19&2.32&3.70&\underline{3.76}&1.75\\
        Contrastive&\textbf{2.49}&\textbf{0.38}&\textbf{0.09}&\textbf{0.97}&\underline{0.488}&\underline{0.89}&\underline{0.28}&\underline{2.86}&\underline{3.72}&\underline{3.76}&\underline{2.07}\\
        MAGIC&\underline{2.51}&\textbf{0.38}&\textbf{0.09}&\textbf{0.97}&\textbf{0.514}&\textbf{0.91}&\textbf{0.36}&\textbf{3.20}$^{\bigstar}$&\textbf{3.76}&\textbf{3.85}&\textbf{2.40}$^{\bigstar}$\\
        \hline
        Human&2.21&0.37&0.09&0.97&0.542&1.00&0.40&3.77&4.11&4.22&2.59\\
		\hlinewd{0.75pt}
	\end{tabular}}
    \caption{Evaluation results on story generation. $\uparrow$ means higher is better and $\downarrow$ means lower is better. The best result is \textbf{bold} and the second best is \underline{underlined}. In human evaluation, ${\bigstar}$ results significantly outperforms the results of other compared methods (Sign Test with p-value < 0.05).}
    	\vspace{-1.5mm}
	\label{tb:story_generation_main_result}
\end{table*}

\subsection{Automatic Evaluation}
\label{sec:story_generation_automatic_evaluation}
Following previous studies \cite{welleck2019neural,meister2022typical,DBLP:journals/corr/abs-2202-06417}, we first evaluate the quality of the generated results from different methods using automatic evaluation metrics, including (1) $n$-gram repetition (rep-$n$); (2) generation diversity (div.); (3) semantic coherence (coh.) between the generated story and the story title; and (4) MAUVE \cite{pillutla2021mauve} score that measures the token distribution closeness between the generated text and the human-written text. In addition, to verify that MAGIC is able to generate stories that are semantically related to the given images, we employ CLIPScore \cite{hessel2021clipscore} to measure the semantic similarity between the generated story and the image retrieved by the story title.

Table \ref{tb:story_generation_main_result} shows the automatic results, from which we observe that MAGIC performs the best on most of the evaluation metrics.\footnote{For stochastic methods (i.e., Top-$k$, Nucleus, and Typical sampling), we report the numbers averaged over 3 runs with different random seeds. We refer to Appendix \ref{appendix:visually_grounded_story_generation} for more details.} The results of rep-$n$, diversity, and MAUVE score demonstrate that MAGIC generates the most diverse stories while being closest to human-written stories in terms of token distribution \cite{pillutla2021mauve}. Moreover, on the coherence (coh.) metric, MAGIC notably outperforms other baselines.  
%Our analysis is 
We conjecture that the image retrieved by the story title contains rich visual concepts and features, therefore providing more grounding information. As a result, by leveraging these visual knowledge, MAGIC can generate stories are more semantically coherent to the story titles. Lastly, on the CLIPScore metric, MAGIC surpasses other methods by large margins, suggesting it generates stories that are more related to the images. In conclusion, the generated text of MAGIC is effectively guided by both the text prompt (i.e., story title) as well as the image, while other methods can only leverage the information from the text prompt.

\subsection{Human Evaluation}
We also conduct a human evaluation with the help of graders proficient in English from a third-party grading platform. We sample 200 instances from the test set. All generated results, plus the reference, are randomly shuffled and evaluated by five graders, resulting in 8,000 annotated samples in total. The evaluation follows a 5-point Likert scale (1, 2, 3, 4, or 5) for each of the following features:\footnote{We refer to Appendix \ref{appendix:evaluation_guideline} for the detailed human evaluation guidelines.}
\begin{itemize}[noitemsep,topsep=1pt]
    \itemsep 0em 
    \item \textbf{Coherence (coh.)}: Whether the generated story is semantically consistent with the title.
    \item \textbf{Fluency (flu.)}: Whether the generated story is fluent and easy to understand.
    \item \textbf{Informativeness (inform.)}: Whether the generated story is diverse and interesting. 
    \item \textbf{Story-Image Relevance (si-rel.)}: Whether the generated story is related to the image that is retrieved by the story title.
\end{itemize}

Table \ref{tb:story_generation_main_result} presents the human evaluation results, with the first row showing strong inter-annotator agreements as measured by Fleiss$'$ kappa coefficient \cite{fleiss1971mns}. %We see that MAGIC attains the best result across the board. Firstly, 
Firstly, on the fluency (flu.) and informativeness (inform.) metrics, MAGIC performs better than other methods. This indicates that the introduction of visual guidance helps the model to generate more interesting content while maintaining the grammaticality and fluency of the generated story. 
%This indicates that the introduction of visual control does not undermine the grammaticality, diversity, as well as interestingness of the generated story. 
Moreover, on the coherence metric, the performance gain of MAGIC over other baselines is significant (Sign Test with p-value < 0.05), showing it better maintains the consistency between the generated story and the story title. This conclusion is also validated by the results of coherence score in the automatic evaluation (Section~\cref{sec:story_generation_automatic_evaluation}). Lastly, on the story-image relevance (si-rel.) metric, MAGIC also outperforms other methods significantly, demonstrating its ability in generating text by visually grounding on the given image.

\begin{figure*}[t] 
  \centering    
  \setlength{\abovecaptionskip}{3pt}
  \includegraphics[width=0.96\textwidth]{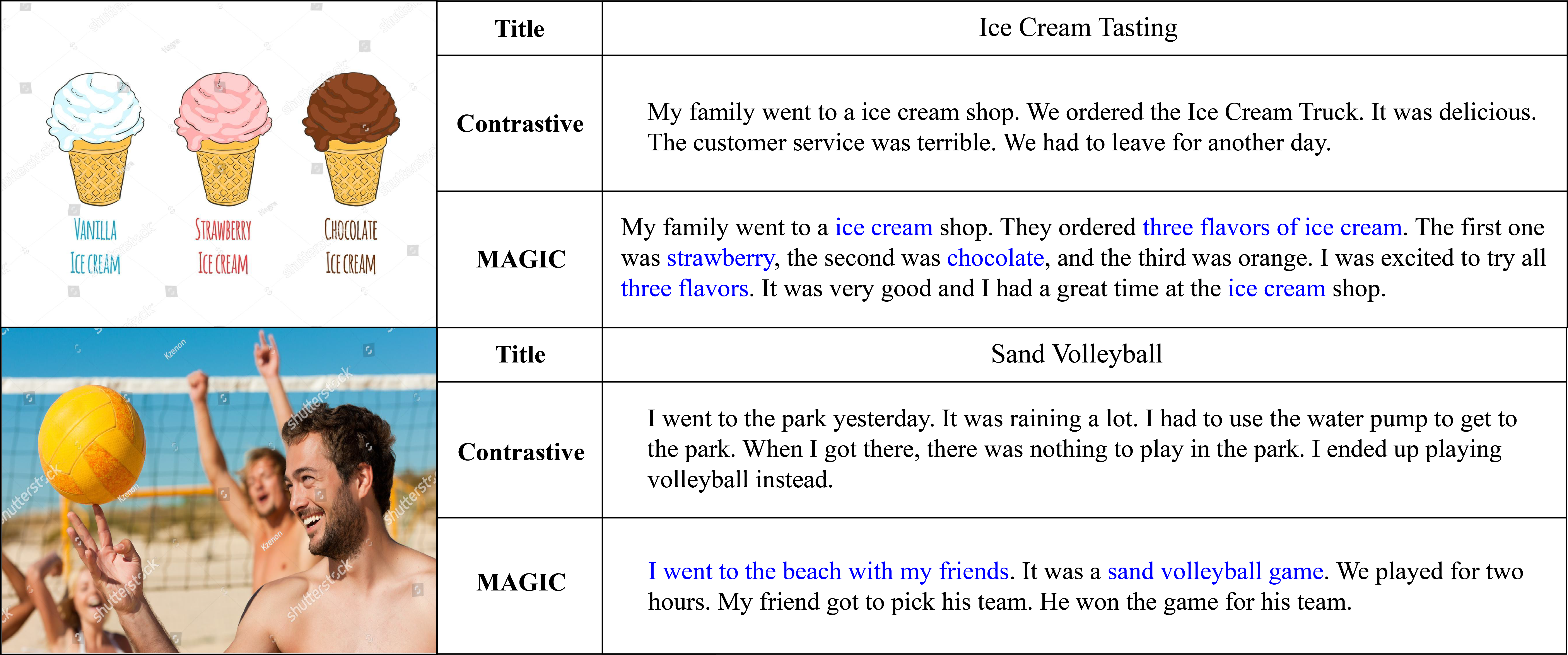}
  \caption{Examples of story generation. MAGIC can generate text (highlighted in \textcolor{blue}{blue}) that is related to the visual concepts displayed in the image. (Best viewed in color and by zooming in.)}
  \label{fig:story_generation_case_study}
  \vspace{-1.5mm}
\end{figure*}

\subsection{Qualitative Evaluation}
In Figure \ref{fig:story_generation_case_study}, we compare our approach with the strongest baseline (i.e., contrastive search), where the image retrieved by the story title is shown on the left-hand side.\footnote{We refer to Appendix \ref{appendix:story_generation} for more examples of story generation.} We see that MAGIC can generate text (highlighted in \textcolor{blue}{blue}) conditioned on the visual concepts of the image. In the first example, MAGIC elaborates details of three types of ice cream. Such details are more interesting as well as more related to the story title (i.e., ice cream tasting) as compared with the story generated by contrastive search. In the second example, the result of contrastive search is clearly off-the-topic. In contrast, by conditioning on the image, the story generated by MAGIC stays on the topic and describes intriguing details about a sand volleyball game: (i) the game was hosted on the beach; (ii) it lasted for two hours; and (iii) the friend won the game. In summary, by leveraging visual guidance from the image, MAGIC can generate semantically coherent story with interesting content.

\section{Conclusion and Future Work}
\label{sec:conclusion_and_future_work}
In this work, we present MAGIC, a novel decoding scheme that plugs visual controls into the generation of a language model. MAGIC is a training-free framework that enables the LM to address challenging multimodal tasks in a zero-shot manner without sacrificing the decoding speed. To verify the versatility and extensibility of MAGIC, we comprehensively evaluate our approach on two image-grounded text generation tasks: (i) image captioning and (ii) visually grounded story generation. Experimental results demonstrate that our approach notably outperforms previous state-of-the-art methods in both automatic and human evaluations. 

\noindent\textbf{Future Work.} While our focus in this study is zero-shot image grounded text generation using a language model, we would like to note that MAGIC Search is a model architecture agnostic decoding scheme. In other words, it can naturally fit into any existing multimodal generative model which takes both the image and text as input. However, it is out of the scope of this paper and we will leave it to future work.

Moreover, in theory, MAGIC is a generic framework that can be extended to modalities beyond text and image. Controls in any form, of any modalities, can be plugged into the language model as long as a certain similarity metric can be found to measure the relevance between the control and the generated text. In future work, we would like to explore the possibility of adapting MAGIC to other modalities beyond images (e.g., audios and videos) therefore enabling the language model to generate text grounded on multimodal intelligence.

\section*{Acknowledgments}
The first author would like to thank Jialu Xu and Qianchu Liu for the insightful discussions and support.

\bibliographystyle{plain}
\bibliography{reference}

\begin{thebibliography}{10}

\bibitem{anderson2016spice}
Peter Anderson, Basura Fernando, Mark Johnson, and Stephen Gould.
\newblock Spice: Semantic propositional image caption evaluation.
\newblock In {\em European Conference on Computer Vision (ECCV)}, 2016.

\bibitem{anderson2018partially}
Peter Anderson, Stephen Gould, and Mark Johnson.
\newblock Partially-supervised image captioning.
\newblock {\em Advances in Neural Information Processing Systems (NeurIPS)},
  2018.

\bibitem{anderson2018bottom}
Peter Anderson, Xiaodong He, Chris Buehler, Damien Teney, Mark Johnson, Stephen
  Gould, and Lei Zhang.
\newblock Bottom-up and top-down attention for image captioning and visual
  question answering.
\newblock In {\em Proceedings of the IEEE/CVF Conference on Computer Vision and
  Pattern Recognition (CVPR)}, 2018.

\bibitem{brown2020language}
Tom Brown, Benjamin Mann, Nick Ryder, Melanie Subbiah, Jared~D Kaplan, Prafulla
  Dhariwal, Arvind Neelakantan, Pranav Shyam, Girish Sastry, Amanda Askell,
  et~al.
\newblock Language models are few-shot learners.
\newblock {\em Advances in Neural Information Processing Systems (NeurIPS)},
  2020.

\bibitem{chatterjee2018diverse}
Moitreya Chatterjee and Alexander~G Schwing.
\newblock Diverse and coherent paragraph generation from images.
\newblock In {\em European Conference on Computer Vision (ECCV)}, 2018.

\bibitem{chen2021visualgpt}
Jun Chen, Han Guo, Kai Yi, Boyang Li, and Mohamed Elhoseiny.
\newblock Visualgpt: Data-efficient adaptation of pretrained language models
  for image captioning.
\newblock {\em arXiv preprint arXiv:2102.10407}, 2021.

\bibitem{chen2021human}
Long Chen, Zhihong Jiang, Jun Xiao, and Wei Liu.
\newblock Human-like controllable image captioning with verb-specific semantic
  roles.
\newblock In {\em Proceedings of the IEEE/CVF Conference on Computer Vision and
  Pattern Recognition (CVPR)}, 2021.

\bibitem{chen2018factual}
Tianlang Chen, Zhongping Zhang, Quanzeng You, Chen Fang, Zhaowen Wang, Hailin
  Jin, and Jiebo Luo.
\newblock ``factual''or``emotional'': Stylized image captioning with adaptive
  learning and attention.
\newblock In {\em European Conference on Computer Vision (ECCV)}, 2018.

\bibitem{chen2020simple}
Ting Chen, Simon Kornblith, Mohammad Norouzi, and Geoffrey Hinton.
\newblock A simple framework for contrastive learning of visual
  representations.
\newblock In {\em International conference on machine learning (ICML)}, 2020.

\bibitem{DBLP:conf/cvpr/ChopraHL05}
Sumit Chopra, Raia Hadsell, and Yann LeCun.
\newblock Learning a similarity metric discriminatively, with application to
  face verification.
\newblock In {\em 2005 {IEEE} Computer Society Conference on Computer Vision
  and Pattern Recognition (CVPR)}, 2005.

\bibitem{chowdhery2022palm}
Aakanksha Chowdhery, Sharan Narang, Jacob Devlin, Maarten Bosma, Gaurav Mishra,
  Adam Roberts, Paul Barham, Hyung~Won Chung, Charles Sutton, Sebastian
  Gehrmann, et~al.
\newblock Palm: Scaling language modeling with pathways.
\newblock {\em arXiv preprint arXiv:2204.02311}, 2022.

\bibitem{DBLP:conf/iclr/ClarkLLM20}
Kevin Clark, Minh{-}Thang Luong, Quoc~V. Le, and Christopher~D. Manning.
\newblock {ELECTRA:} pre-training text encoders as discriminators rather than
  generators.
\newblock In {\em 8th International Conference on Learning Representations,
  (ICLR)}, 2020.

\bibitem{das2021container}
Sarkar Snigdha~Sarathi Das, Arzoo Katiyar, Rebecca~J Passonneau, and Rui Zhang.
\newblock Container: Few-shot named entity recognition via contrastive
  learning.
\newblock {\em arXiv preprint arXiv:2109.07589}, 2021.

\bibitem{dathathri2020plug}
Sumanth Dathathri, Andrea Madotto, Janice Lan, Jane Hung, Eric Frank, Piero
  Molino, Jason Yosinski, and Rosanne Liu.
\newblock Plug and play language models: A simple approach to controlled text
  generation.
\newblock In {\em International Conference on Learning Representations (ICLR)},
  2020.

\bibitem{denkowski2014meteor}
Michael Denkowski and Alon Lavie.
\newblock Meteor universal: Language specific translation evaluation for any
  target language.
\newblock In {\em Proceedings of the Workshop on Statistical Machine
  Translation}, 2014.

\bibitem{DBLP:conf/naacl/DevlinCLT19}
Jacob Devlin, Ming{-}Wei Chang, Kenton Lee, and Kristina Toutanova.
\newblock {BERT:} pre-training of deep bidirectional transformers for language
  understanding.
\newblock In {\em Proceedings of the 2019 Conference of the North American
  Chapter of the Association for Computational Linguistics: Human Language
  Technologies (NAACL)}, 2019.

\bibitem{dosovitskiy2021image}
Alexey Dosovitskiy, Lucas Beyer, Alexander Kolesnikov, Dirk Weissenborn,
  Xiaohua Zhai, Thomas Unterthiner, Mostafa Dehghani, Matthias Minderer, Georg
  Heigold, Sylvain Gelly, et~al.
\newblock An image is worth 16x16 words: Transformers for image recognition at
  scale.
\newblock In {\em International Conference on Learning Representations (ICLR)},
  2021.

\bibitem{DBLP:conf/acl/LewisDF18}
Angela Fan, Mike Lewis, and Yann~N. Dauphin.
\newblock Hierarchical neural story generation.
\newblock In Iryna Gurevych and Yusuke Miyao, editors, {\em Proceedings of the
  Annual Meeting of the Association for Computational Linguistics (ACL)}, 2018.

\bibitem{feng2019unsupervised}
Yang Feng, Lin Ma, Wei Liu, and Jiebo Luo.
\newblock Unsupervised image captioning.
\newblock In {\em Proceedings of the IEEE/CVF Conference on Computer Vision and
  Pattern Recognition (CVPR)}, 2019.

\bibitem{fleiss1971mns}
J.L. Fleiss et~al.
\newblock {Measuring nominal scale agreement among many raters}.
\newblock {\em Psychological Bulletin}, 76(5):378--382, 1971.

\bibitem{gan2017stylenet}
Chuang Gan, Zhe Gan, Xiaodong He, Jianfeng Gao, and Li~Deng.
\newblock Stylenet: Generating attractive visual captions with styles.
\newblock In {\em Proceedings of the IEEE/CVF Conference on Computer Vision and
  Pattern Recognition (CVPR)}, 2017.

\bibitem{gao2021simcse}
Tianyu Gao, Xingcheng Yao, and Danqi Chen.
\newblock {S}im{CSE}: Simple contrastive learning of sentence embeddings.
\newblock In {\em Proceedings of the 2021 Conference on Empirical Methods in
  Natural Language Processing (EMNLP)}, 2021.

\bibitem{hessel2021clipscore}
Jack Hessel, Ari Holtzman, Maxwell Forbes, Ronan~Le Bras, and Yejin Choi.
\newblock Clipscore: A reference-free evaluation metric for image captioning.
\newblock {\em arXiv preprint arXiv:2104.08718}, 2021.

\bibitem{DBLP:conf/iclr/HoltzmanBDFC20}
Ari Holtzman, Jan Buys, Li~Du, Maxwell Forbes, and Yejin Choi.
\newblock The curious case of neural text degeneration.
\newblock In {\em International Conference on Learning Representations (ICLR)},
  2020.

\bibitem{Honda2021RemovingWS}
Ukyo Honda, Y.~Ushiku, Atsushi Hashimoto, Taro Watanabe, and Yuji Matsumoto.
\newblock Removing word-level spurious alignment between images and
  pseudo-captions in unsupervised image captioning.
\newblock In {\em Proceedings of the Conference of the European Chapter of the
  Association for Computational Linguistics (EACL)}, 2021.

\bibitem{hu2021scaling}
Xiaowei Hu, Zhe Gan, Jianfeng Wang, Zhengyuan Yang, Zicheng Liu, Yumao Lu, and
  Lijuan Wang.
\newblock Scaling up vision-language pre-training for image captioning.
\newblock {\em arXiv preprint arXiv:2111.12233}, 2021.

\bibitem{huang2021unifying}
Yupan Huang, Hongwei Xue, Bei Liu, and Yutong Lu.
\newblock Unifying multimodal transformer for bi-directional image and text
  generation.
\newblock In {\em Proceedings of the 29th ACM International Conference on
  Multimedia}, pages 1138--1147, 2021.

\bibitem{jia2021scaling}
Chao Jia, Yinfei Yang, Ye~Xia, Yi-Ting Chen, Zarana Parekh, Hieu Pham, Quoc Le,
  Yun-Hsuan Sung, Zhen Li, and Tom Duerig.
\newblock Scaling up visual and vision-language representation learning with
  noisy text supervision.
\newblock In {\em International Conference on Machine Learning (ICML)}, 2021.

\bibitem{johnson2019billion}
Jeff Johnson, Matthijs Douze, and Herv{\'e} J{\'e}gou.
\newblock Billion-scale similarity search with gpus.
\newblock {\em IEEE Transactions on Big Data}, 7(3):535--547, 2019.

\bibitem{johnson2016densecap}
Justin Johnson, Andrej Karpathy, and Li~Fei-Fei.
\newblock Densecap: Fully convolutional localization networks for dense
  captioning.
\newblock In {\em Proceedings of the IEEE/CVF Conference on Computer Vision and
  Pattern Recognition (CVPR)}, 2016.

\bibitem{karpathy2015deep}
Andrej Karpathy and Li~Fei-Fei.
\newblock Deep visual-semantic alignments for generating image descriptions.
\newblock In {\em Proceedings of the IEEE/CVF Conference on Computer Vision and
  Pattern Recognition (CVPR)}, 2015.

\bibitem{karras2021alias}
Tero Karras, Miika Aittala, Samuli Laine, Erik H{\"a}rk{\"o}nen, Janne
  Hellsten, Jaakko Lehtinen, and Timo Aila.
\newblock Alias-free generative adversarial networks.
\newblock {\em Advances in Neural Information Processing Systems (NeurIPS)},
  2021.

\bibitem{karras2019style}
Tero Karras, Samuli Laine, and Timo Aila.
\newblock A style-based generator architecture for generative adversarial
  networks.
\newblock In {\em Proceedings of the IEEE/CVF Conference on Computer Vision and
  Pattern Recognition (CVPR)}, 2019.

\bibitem{karras2020analyzing}
Tero Karras, Samuli Laine, Miika Aittala, Janne Hellsten, Jaakko Lehtinen, and
  Timo Aila.
\newblock Analyzing and improving the image quality of stylegan.
\newblock In {\em Proceedings of the IEEE/CVF Conference on Computer Vision and
  Pattern Recognition (CVPR)}, 2020.

\bibitem{kim2019dense}
Dong-Jin Kim, Jinsoo Choi, Tae-Hyun Oh, and In~So Kweon.
\newblock Dense relational captioning: Triple-stream networks for
  relationship-based captioning.
\newblock In {\em Proceedings of the IEEE/CVF Conference on Computer Vision and
  Pattern Recognition (CVPR)}, 2019.

\bibitem{DBLP:journals/corr/KingmaB14}
Diederik~P. Kingma and Jimmy Ba.
\newblock Adam: {A} method for stochastic optimization.
\newblock In Yoshua Bengio and Yann LeCun, editors, {\em International
  Conference on Learning Representations (ICLR)}, 2015.

\bibitem{Laina2019TowardsUI}
Iro Laina, C.~Rupprecht, and Nassir Navab.
\newblock Towards unsupervised image captioning with shared multimodal
  embeddings.
\newblock {\em Proceedings of the IEEE/CVF International Conference on Computer
  Vision (ICCV)}, 2019.

\bibitem{DBLP:journals/corr/abs-2110-06612}
Tian Lan, Deng Cai, Yan Wang, Yixuan Su, Xian-Ling Mao, and Heyan Huang.
\newblock Exploring dense retrieval for dialogue response selection.
\newblock {\em arXiv preprint arXiv:2110.06612}, 2021.

\bibitem{DBLP:conf/acl/LewisLGGMLSZ20}
Mike Lewis, Yinhan Liu, Naman Goyal, Marjan Ghazvininejad, Abdelrahman Mohamed,
  Omer Levy, Veselin Stoyanov, and Luke Zettlemoyer.
\newblock {BART:} denoising sequence-to-sequence pre-training for natural
  language generation, translation, and comprehension.
\newblock In {\em Proceedings of the Annual Meeting of the Association for
  Computational Linguistics (ACL)}, 2020.

\bibitem{Li2020OscarOA}
Xiujun Li, Xi~Yin, Chunyuan Li, Xiaowei Hu, Pengchuan Zhang, Lei Zhang, Lijuan
  Wang, Houdong Hu, Li~Dong, Furu Wei, Yejin Choi, and Jianfeng Gao.
\newblock Oscar: Object-semantics aligned pre-training for vision-language
  tasks.
\newblock In {\em European Conference on Computer Vision (ECCV)}, 2020.

\bibitem{lin2004automatic}
Chin-Yew Lin and Franz~Josef Och.
\newblock Automatic evaluation of machine translation quality using longest
  common subsequence and skip-bigram statistics.
\newblock In {\em Proceedings of the Annual Meeting of the Association for
  Computational Linguistics (ACL)}, 2004.

\bibitem{lin2014microsoft}
Tsung-Yi Lin, Michael Maire, Serge Belongie, James Hays, Pietro Perona, Deva
  Ramanan, Piotr Doll{\'a}r, and C~Lawrence Zitnick.
\newblock Microsoft coco: Common objects in context.
\newblock In {\em European Conference on Computer Vision (ECCV)}, 2014.

\bibitem{liu2021fast}
Fangyu Liu, Ivan Vuli{\'c}, Anna Korhonen, and Nigel Collier.
\newblock Fast, effective, and self-supervised: Transforming masked language
  models into universal lexical and sentence encoders.
\newblock In {\em Proceedings of the 2021 Conference on Empirical Methods in
  Natural Language Processing (EMNLP)}, 2021.

\bibitem{DBLP:journals/corr/abs-1907-11692}
Yinhan Liu, Myle Ott, Naman Goyal, Jingfei Du, Mandar Joshi, Danqi Chen, Omer
  Levy, Mike Lewis, Luke Zettlemoyer, and Veselin Stoyanov.
\newblock Roberta: {A} robustly optimized {BERT} pretraining approach.
\newblock {\em CoRR}, abs/1907.11692, 2019.

\bibitem{liu-liu-2021-simcls}
Yixin Liu and Pengfei Liu.
\newblock {S}im{CLS}: A simple framework for contrastive learning of
  abstractive summarization.
\newblock In {\em Proceedings of the 59th Annual Meeting of the Association for
  Computational Linguistics and the 11th International Joint Conference on
  Natural Language Processing (ACL)}, 2021.

\bibitem{lu2017knowing}
Jiasen Lu, Caiming Xiong, Devi Parikh, and Richard Socher.
\newblock Knowing when to look: Adaptive attention via a visual sentinel for
  image captioning.
\newblock In {\em Proceedings of the IEEE/CVF Conference on Computer Vision and
  Pattern Recognition (CVPR)}, 2017.

\bibitem{mao2014explain}
Junhua Mao, Wei Xu, Yi~Yang, Jiang Wang, and Alan~L Yuille.
\newblock Explain images with multimodal recurrent neural networks.
\newblock {\em arXiv preprint arXiv:1410.1090}, 2014.

\bibitem{mathews2016senticap}
Alexander Mathews, Lexing Xie, and Xuming He.
\newblock Senticap: Generating image descriptions with sentiments.
\newblock In {\em Proceedings of the AAAI conference on artificial intelligence
  (AAAI)}, 2016.

\bibitem{meister2022typical}
Clara Meister, Tiago Pimentel, Gian Wiher, and Ryan Cotterell.
\newblock Typical decoding for natural language generation.
\newblock {\em arXiv preprint arXiv:2202.00666}, 2022.

\bibitem{DBLP:journals/corr/abs-2110-08173}
Zaiqiao Meng, Fangyu Liu, Ehsan Shareghi, Yixuan Su, Charlotte Collins, and
  Nigel Collier.
\newblock Rewire-then-probe: {A} contrastive recipe for probing biomedical
  knowledge of pre-trained language models.
\newblock {\em CoRR}, abs/2110.08173, 2021.

\bibitem{Mokady2021ClipCapCP}
Ron Mokady, Amir Hertz, and Amit~H. Bermano.
\newblock Clipcap: Clip prefix for image captioning.
\newblock {\em ArXiv}, abs/2111.09734, 2021.

\bibitem{mostafazadeh2016corpus}
Nasrin Mostafazadeh, Nathanael Chambers, Xiaodong He, Devi Parikh, Dhruv Batra,
  Lucy Vanderwende, Pushmeet Kohli, and James Allen.
\newblock A corpus and cloze evaluation for deeper understanding of commonsense
  stories.
\newblock In {\em Proceedings of the Conference of the North American Chapter
  of the Association for Computational Linguistics (NAACL)}, 2016.

\bibitem{nguyen2017plug}
Anh Nguyen, Jeff Clune, Yoshua Bengio, Alexey Dosovitskiy, and Jason Yosinski.
\newblock Plug \& play generative networks: Conditional iterative generation of
  images in latent space.
\newblock In {\em Proceedings of the IEEE/CVF Conference on Computer Vision and
  Pattern Recognition (CVPR}, 2017.

\bibitem{nguyen2016synthesizing}
Anh Nguyen, Alexey Dosovitskiy, Jason Yosinski, Thomas Brox, and Jeff Clune.
\newblock Synthesizing the preferred inputs for neurons in neural networks via
  deep generator networks.
\newblock {\em Advances in Neural Information Processing Systems (NeurIPS)},
  2016.

\bibitem{papineni2002bleu}
Kishore Papineni, Salim Roukos, Todd Ward, and Wei-Jing Zhu.
\newblock Bleu: a method for automatic evaluation of machine translation.
\newblock In {\em Proceedings of the Annual Meeting of the Association for
  Computational Linguistics (ACL)}, 2002.

\bibitem{patashnik2021styleclip}
Or~Patashnik, Zongze Wu, Eli Shechtman, Daniel Cohen-Or, and Dani Lischinski.
\newblock Styleclip: Text-driven manipulation of stylegan imagery.
\newblock In {\em Proceedings of the IEEE/CVF International Conference on
  Computer Vision (ICCV)}, 2021.

\bibitem{pillutla2021mauve}
Krishna Pillutla, Swabha Swayamdipta, Rowan Zellers, John Thickstun, Sean
  Welleck, Yejin Choi, and Zaid Harchaoui.
\newblock Mauve: Measuring the gap between neural text and human text using
  divergence frontiers.
\newblock {\em Advances in Neural Information Processing Systems (NeurIPS)},
  2021.

\bibitem{plummer2015flickr30k}
Bryan~A Plummer, Liwei Wang, Chris~M Cervantes, Juan~C Caicedo, Julia
  Hockenmaier, and Svetlana Lazebnik.
\newblock Flickr30k entities: Collecting region-to-phrase correspondences for
  richer image-to-sentence models.
\newblock In {\em Proceedings of the IEEE/CVF International Conference on
  Computer Vision (ICCV)}, 2015.

\bibitem{radford2021learning}
Alec Radford, Jong~Wook Kim, Chris Hallacy, Aditya Ramesh, Gabriel Goh,
  Sandhini Agarwal, Girish Sastry, Amanda Askell, Pamela Mishkin, Jack Clark,
  et~al.
\newblock Learning transferable visual models from natural language
  supervision.
\newblock In {\em International Conference on Machine Learning (ICML)}, 2021.

\bibitem{radford2019language}
Alec Radford, Jeff Wu, Rewon Child, David Luan, Dario Amodei, and Ilya
  Sutskever.
\newblock Language models are unsupervised multitask learners.
\newblock 2019.

\bibitem{DBLP:journals/jmlr/RaffelSRLNMZLL20}
Colin Raffel, Noam Shazeer, Adam Roberts, Katherine Lee, Sharan Narang, Michael
  Matena, Yanqi Zhou, Wei Li, and Peter~J Liu.
\newblock Exploring the limits of transfer learning with a unified text-to-text
  transformer.
\newblock {\em Journal of Machine Learning Research}, 21:1--67, 2020.

\bibitem{ramesh2022hierarchical}
Aditya Ramesh, Prafulla Dhariwal, Alex Nichol, Casey Chu, and Mark Chen.
\newblock Hierarchical text-conditional image generation with clip latents.
\newblock {\em arXiv preprint arXiv:2204.06125}, 2022.

\bibitem{ramesh2021zero}
Aditya Ramesh, Mikhail Pavlov, Gabriel Goh, Scott Gray, Chelsea Voss, Alec
  Radford, Mark Chen, and Ilya Sutskever.
\newblock Zero-shot text-to-image generation.
\newblock In {\em International Conference on Machine Learning (ICLR)}, 2021.

\bibitem{roberts1998optimal}
Gareth~O Roberts and Jeffrey~S Rosenthal.
\newblock Optimal scaling of discrete approximations to langevin diffusions.
\newblock {\em Journal of the Royal Statistical Society: Series B (Statistical
  Methodology)}, 60(1):255--268, 1998.

\bibitem{roberts1996exponential}
Gareth~O Roberts and Richard~L Tweedie.
\newblock Exponential convergence of langevin distributions and their discrete
  approximations.
\newblock {\em Bernoulli}, pages 341--363, 1996.

\bibitem{DBLP:conf/icra/SermanetLCHJSLB18}
Pierre Sermanet, Corey Lynch, Yevgen Chebotar, Jasmine Hsu, Eric Jang, Stefan
  Schaal, and Sergey Levine.
\newblock Time-contrastive networks: Self-supervised learning from video.
\newblock In {\em 2018 {IEEE} International Conference on Robotics and
  Automation (ICRA)}, 2018.

\bibitem{sharma2018conceptual}
Piyush Sharma, Nan Ding, Sebastian Goodman, and Radu Soricut.
\newblock Conceptual captions: A cleaned, hypernymed, image alt-text dataset
  for automatic image captioning.
\newblock In {\em Proceedings of the Annual Meeting of the Association for
  Computational Linguistics (ACL)}, 2018.

\bibitem{shen2021much}
Sheng Shen, Liunian~Harold Li, Hao Tan, Mohit Bansal, Anna Rohrbach, Kai-Wei
  Chang, Zhewei Yao, and Kurt Keutzer.
\newblock How much can clip benefit vision-and-language tasks?
\newblock {\em arXiv preprint arXiv:2107.06383}, 2021.

\bibitem{shen2020interpreting}
Yujun Shen, Jinjin Gu, Xiaoou Tang, and Bolei Zhou.
\newblock Interpreting the latent space of gans for semantic face editing.
\newblock In {\em Proceedings of the IEEE/CVF Conference on Computer Vision and
  Pattern Recognition (CVPR)}, 2020.

\bibitem{shuster2019engaging}
Kurt Shuster, Samuel Humeau, Hexiang Hu, Antoine Bordes, and Jason Weston.
\newblock Engaging image captioning via personality.
\newblock In {\em Proceedings of the IEEE/CVF Conference on Computer Vision and
  Pattern Recognition (CVPR)}, 2019.

\bibitem{DBLP:conf/eacl/SuCWVBLC21}
Yixuan Su, Deng Cai, Yan Wang, David Vandyke, Simon Baker, Piji Li, and Nigel
  Collier.
\newblock Non-autoregressive text generation with pre-trained language models.
\newblock In {\em Proceedings of the 16th Conference of the European Chapter of
  the Association for Computational Linguistics: Main Volume (EACL)}, 2021.

\bibitem{DBLP:journals/corr/abs-2202-06417}
Yixuan Su, Tian Lan, Yan Wang, Dani Yogatama, Lingpeng Kong, and Nigel Collier.
\newblock A contrastive framework for neural text generation.
\newblock {\em CoRR}, abs/2202.06417, 2022.

\bibitem{DBLP:journals/corr/abs-2111-04198}
Yixuan Su, Fangyu Liu, Zaiqiao Meng, Tian Lan, Lei Shu, Ehsan Shareghi, and
  Nigel Collier.
\newblock Tacl: Improving {BERT} pre-training with token-aware contrastive
  learning.
\newblock {\em CoRR}, abs/2111.04198, 2021.

\bibitem{DBLP:conf/emnlp/SuMBC21}
Yixuan Su, Zaiqiao Meng, Simon Baker, and Nigel Collier.
\newblock Few-shot table-to-text generation with prototype memory.
\newblock In Marie{-}Francine Moens, Xuanjing Huang, Lucia Specia, and
  Scott~Wen{-}tau Yih, editors, {\em Findings of the Association for
  Computational Linguistics: {EMNLP}}, 2021.

\bibitem{DBLP:journals/corr/abs-2109-14739}
Yixuan Su, Lei Shu, Elman Mansimov, Arshit Gupta, Deng Cai, Yi{-}An Lai, and
  Yi~Zhang.
\newblock Multi-task pre-training for plug-and-play task-oriented dialogue
  system.
\newblock {\em CoRR}, abs/2109.14739, 2021.

\bibitem{DBLP:conf/emnlp/SuVWFC21}
Yixuan Su, David Vandyke, Sihui Wang, Yimai Fang, and Nigel Collier.
\newblock Plan-then-generate: Controlled data-to-text generation via planning.
\newblock In Marie{-}Francine Moens, Xuanjing Huang, Lucia Specia, and
  Scott~Wen{-}tau Yih, editors, {\em Findings of the Association for
  Computational Linguistics: {EMNLP}}, 2021.

\bibitem{DBLP:journals/taslp/SuWCBKC21}
Yixuan Su, Yan Wang, Deng Cai, Simon Baker, Anna Korhonen, and Nigel Collier.
\newblock {PROTOTYPE-TO-STYLE:} dialogue generation with style-aware editing on
  retrieval memory.
\newblock {\em {IEEE} {ACM} Trans. Audio Speech Lang. Process.}, 29:2152--2161,
  2021.

\bibitem{tewel2021zero}
Yoad Tewel, Yoav Shalev, Idan Schwartz, and Lior Wolf.
\newblock Zero-shot image-to-text generation for visual-semantic arithmetic.
\newblock {\em arXiv preprint arXiv:2111.14447}, 2021.

\bibitem{vaswani2017attention}
Ashish Vaswani, Noam Shazeer, Niki Parmar, Jakob Uszkoreit, Llion Jones,
  Aidan~N. Gomez, Lukasz Kaiser, and Illia Polosukhin.
\newblock Attention is all you need.
\newblock In {\em Advances in Neural Information Processing Systems (NeurIPS)},
  2017.

\bibitem{vedantam2015cider}
Ramakrishna Vedantam, C~Lawrence~Zitnick, and Devi Parikh.
\newblock Cider: Consensus-based image description evaluation.
\newblock In {\em Proceedings of the IEEE/CVF Conference on Computer Vision and
  Pattern Recognition (CVPR)}, 2015.

\bibitem{vinyals2015show}
Oriol Vinyals, Alexander Toshev, Samy Bengio, and Dumitru Erhan.
\newblock Show and tell: A neural image caption generator.
\newblock In {\em Proceedings of the IEEE/CVF Conference on Computer Vision and
  Pattern Recognition (CVPR)}, 2015.

\bibitem{DBLP:conf/iccv/WangG15}
Xiaolong Wang and Abhinav Gupta.
\newblock Unsupervised learning of visual representations using videos.
\newblock In {\em 2015 {IEEE} International Conference on Computer Vision
  (ICCV)}, 2015.

\bibitem{welleck2019neural}
Sean Welleck, Ilia Kulikov, Stephen Roller, Emily Dinan, Kyunghyun Cho, and
  Jason Weston.
\newblock Neural text generation with unlikelihood training.
\newblock {\em arXiv preprint arXiv:1908.04319}, 2019.

\bibitem{xu2021videoclip}
Hu~Xu, Gargi Ghosh, Po{-}Yao Huang, Dmytro Okhonko, Armen Aghajanyan, Florian
  Metze, Luke Zettlemoyer, and Christoph Feichtenhofer.
\newblock Videoclip: Contrastive pre-training for zero-shot video-text
  understanding.
\newblock In {\em Proceedings of the 2021 Conference on Empirical Methods in
  Natural Language Processing (EMNLP)}, 2021.

\bibitem{xu2015show}
Kelvin Xu, Jimmy Ba, Ryan Kiros, Kyunghyun Cho, Aaron Courville, Ruslan
  Salakhudinov, Rich Zemel, and Yoshua Bengio.
\newblock Show, attend and tell: Neural image caption generation with visual
  attention.
\newblock In {\em International Conference on Machine Learning (ICML)}, 2015.

\bibitem{yang2021taco}
Jianwei Yang, Yonatan Bisk, and Jianfeng Gao.
\newblock Taco: Token-aware cascade contrastive learning for video-text
  alignment.
\newblock In {\em Proceedings of the IEEE/CVF International Conference on
  Computer Vision (ICCV)}, 2021.

\bibitem{DBLP:conf/nips/YangDYCSL19}
Zhilin Yang, Zihang Dai, Yiming Yang, Jaime~G. Carbonell, Ruslan Salakhutdinov,
  and Quoc~V. Le.
\newblock Xlnet: Generalized autoregressive pretraining for language
  understanding.
\newblock In {\em Advances in Neural Information Processing Systems (NeurIPS)},
  2019.

\bibitem{yin2019context}
Guojun Yin, Lu~Sheng, Bin Liu, Nenghai Yu, Xiaogang Wang, and Jing Shao.
\newblock Context and attribute grounded dense captioning.
\newblock In {\em Proceedings of the IEEE/CVF Conference on Computer Vision and
  Pattern Recognition (CVPR)}, 2019.

\bibitem{zeng2020dense}
Runhao Zeng, Haoming Xu, Wenbing Huang, Peihao Chen, Mingkui Tan, and Chuang
  Gan.
\newblock Dense regression network for video grounding.
\newblock In {\em Proceedings of the IEEE/CVF Conference on Computer Vision and
  Pattern Recognition (CVPR)}, 2020.

\bibitem{zhang2021vinvl}
Pengchuan Zhang, Xiujun Li, Xiaowei Hu, Jianwei Yang, Lei Zhang, Lijuan Wang,
  Yejin Choi, and Jianfeng Gao.
\newblock Vinvl: Revisiting visual representations in vision-language models.
\newblock In {\em Proceedings of the IEEE/CVF Conference on Computer Vision and
  Pattern Recognition (CVPR)}, 2021.

\bibitem{zhang2021rstnet}
Xuying Zhang, Xiaoshuai Sun, Yunpeng Luo, Jiayi Ji, Yiyi Zhou, Yongjian Wu,
  Feiyue Huang, and Rongrong Ji.
\newblock Rstnet: Captioning with adaptive attention on visual and non-visual
  words.
\newblock In {\em Proceedings of the IEEE/CVF Conference on Computer Vision and
  Pattern Recognition (CVPR)}, 2021.

\bibitem{DBLP:journals/corr/abs-2109-02492}
Ming Zhong, Yang Liu, Yichong Xu, Chenguang Zhu, and Michael Zeng.
\newblock Dialoglm: Pre-trained model for long dialogue understanding and
  summarization.
\newblock {\em CoRR}, abs/2109.02492, 2021.

\bibitem{zhou2019grounded}
Luowei Zhou, Yannis Kalantidis, Xinlei Chen, Jason~J Corso, and Marcus
  Rohrbach.
\newblock Grounded video description.
\newblock In {\em Proceedings of the IEEE/CVF Conference on Computer Vision and
  Pattern Recognition (CVPR)}, 2019.

\bibitem{zhou2020unified}
Luowei Zhou, Hamid Palangi, Lei Zhang, Houdong Hu, Jason Corso, and Jianfeng
  Gao.
\newblock Unified vision-language pre-training for image captioning and vqa.
\newblock In {\em Proceedings of the AAAI Conference on Artificial Intelligence
  (AAAI)}, 2020.

\end{thebibliography}

%%%% BEGIN INSTRUCTIONS %%%
%%The checklist follows the references.  Please
%%read the checklist guidelines carefully for information on how to answer %these
%%questions.  For each question, change the default \answerTODO{} to %%\answerYes{},
%%\answerNo{}, or \answerNA{}.  You are strongly encouraged to include a {\bf
%%justification to your answer}, either by referencing the appropriate %section %of
%%your paper or providing a brief inline description.  For example:
%%\begin{itemize}
%%  \item Did you include the license to the code and datasets? %\answerYes{See %Section~\ref{gen_inst}.}
%%  \item Did you include the license to the code and datasets? \answerNo{The %%code and the data are proprietary.}
%%  \item Did you include the license to the code and datasets? \answerNA{}
%%\end{itemize}
%%Please do not modify the questions and only use the provided macros for %your
%%answers.  Note that the Checklist section does not count towards the page
%%limit.  In your paper, please delete this instructions block and only keep %the
%%Checklist section heading above along with the questions/answers below.

%\clearpage
%%%%%%%%%%%%%%%%%%%%%%%%%%%%%%%%%%%%%%%%%%%%%%%%%%%%%%%%%%%%%%%%%%%%%%%%%%%%%%%%%%%%%%%%%%%%%%%%%%%%%%%%%%%%%%%%%%%%%%%%%%%%%%%%%%%%%%%%%%%%%%%%%%%%%%%%%%%%

% Checklist Location

\clearpage

%%%%%%%%%%%%%%%%%%%%%%%%%%%%%%%%%%%%%%%%%%%%%%%%%%%%%%%%%%%%%%%%%%%%%%%%%%%%%%%%%%%%%%%%%%%%%%%%%%%%%%%%%%%%%%%%%%%%%%%%%%%%%%%%%%%%%%%%%%%%%%%%%%%%%%%%%%%%

\clearpage

\appendix
\addcontentsline{toc}{section}{Appendix} % Add the appendix text to the document TOC
\part{Appendix} % Start the appendix part
\parttoc % Insert the appendix TOC
\clearpage

\section{Related Work}
In this section, we describe other researches that are related to our work.

\textbf{Pre-trained Language Models.} Since the rising of GPT-2~\cite{radford2019language} and  BERT~\cite{DBLP:conf/naacl/DevlinCLT19}, the research community has witnessed remarkable progress in the field of language model pre-training on a large amount of free text. Such advancements have led to significant progresses in a wide range of natural language understanding (NLU) tasks \cite{DBLP:journals/corr/abs-1907-11692,DBLP:conf/nips/YangDYCSL19,DBLP:conf/iclr/ClarkLLM20,DBLP:journals/corr/abs-2110-06612} and text generation tasks tasks~\cite{radford2019language,DBLP:conf/acl/LewisLGGMLSZ20,DBLP:journals/jmlr/RaffelSRLNMZLL20,DBLP:conf/eacl/SuCWVBLC21,DBLP:conf/emnlp/SuVWFC21,DBLP:journals/corr/abs-2109-14739,DBLP:journals/corr/abs-2109-02492,DBLP:conf/emnlp/SuMBC21,DBLP:journals/taslp/SuWCBKC21}.

\textbf{Contrastive Learning.} Generally, contrastive learning methods distinguish observed data points from fictitious negative samples. They have been widely applied to various computer vision areas, including image \cite{DBLP:conf/cvpr/ChopraHL05} and video \cite{DBLP:conf/iccv/WangG15,DBLP:conf/icra/SermanetLCHJSLB18}. Recently, Chen \emph{et al.}~\cite{chen2020simple} propose a simple framework for contrastive learning of visual representations (SimCLR) based on multi-class N-pair loss. Radford \emph{et al.}~\cite{radford2021learning} and Jia \emph{et al.}~\cite{jia2021scaling} apply the contrastive learning approach for language-image pretraining. Xu \emph{et al.}~\cite{xu2021videoclip} and Yang \emph{et al.}~\cite{yang2021taco} propose a contrastive pre-training approach for video-text alignment.

In the field of NLP, numerous approaches have been proposed to learn better token-level~\cite{DBLP:journals/corr/abs-2111-04198} and sentence-level~\cite{gao2021simcse,liu2021fast} representations using contrastive learning. Beyond representation learning, contrastive learning has also been applied to other NLP applications such as name entity recognition (NER)~\cite{das2021container} and document summarisation \cite{liu-liu-2021-simcls}, knowledge probing for pre-trained language models \cite{DBLP:journals/corr/abs-2110-08173}, and open-ended text generation \cite{DBLP:journals/corr/abs-2202-06417}.

\section{Ablation Study on Hyperparameters of MAGIC Search}
\label{appendix:ablation_study}

\begin{figure*}[hb] 
  \centering    
  \setlength{\abovecaptionskip}{3pt}
  \includegraphics[width=1.0\textwidth]{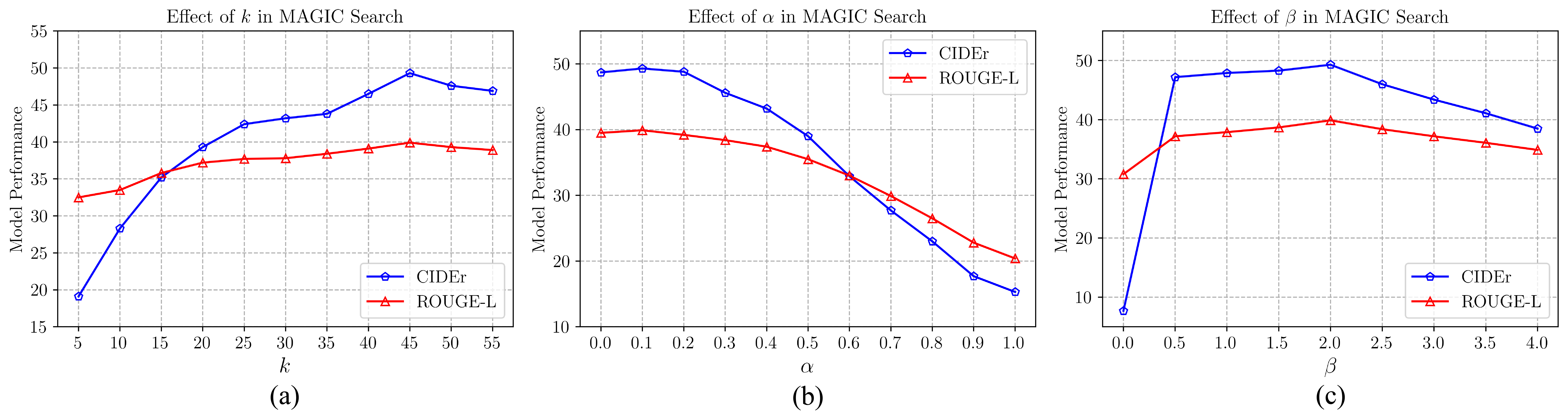}
  \caption{Ablation Study on MS-COCO: (a) Effect of $k$ on MAGIC Search; (b) Effect of $\alpha$ on MAGIC Search; (c) Effect of $\beta$ on MAGIC Search. (Best viewed in color and by zooming in.)}
  \label{fig:magic_ablation_study}
  \vspace{-1.5mm}
\end{figure*}

In this section, we provide further analysis on the effect of hyperparameters in MAGIC Search. To this end, we conduct ablation study experiments on MS-COCO. Recall from Section \cref{sec:image_captioning} that we set $k$, $\alpha$, and $\beta$ in Eq. (\ref{eq:score}) for MS-COCO as 45, 0.1, 2.0, respectively. To isolate the effect of each hyperparameter, in every experiment, we only vary the value of one hyperparamter while keeping others constant. 

\textbf{Effect of $k$.} Figure \ref{fig:magic_ablation_study}(a) shows the  performances (i.e., CIDEr and ROUGE-L) of MAGIC Search by varing $k$ from 5 to 55. We observe that, when $k$ is too small (i.e., $k\leq40$), the performances  are not optimal. The reason is that, a small $k$ leads to a too constrained search space, therefore MAGIC Search cannot find the optimal text sequence that best describes the given image. On the other hand, when $k$ is too large (i.e., $k\geq50$), the search space becomes too large therefore introducing extra noise that causes the performance drop in CIDEr and ROUGE-L. In our experiments, the optimal setup for $k$ in MAGIC Search is 45.

\textbf{Effect of $\alpha$.} Figure \ref{fig:magic_ablation_study}(b) demonstrates the performances on MS-COCO by varying $\alpha$ from 0.0 to 1.0. We see that, when $\alpha$ is small (i.e., $\alpha\leq0.2$), the performances are relatively the same. On the other hand, a large $\alpha$ (i.e., $\alpha\geq0.3$) causes notable drop in the performances. This is due to the fact that a large $\alpha$ forces the language model to generate the text continuation that is less semantically similar to the previously generated context~\cite{DBLP:journals/corr/abs-2202-06417}, therefore affecting the performances of MAGIC Search. In our experiments, the optimal setup for $\alpha$ is 0.1.

\textbf{Effect of $\beta$.} Lastly, Figure \ref{fig:magic_ablation_study}(c) illustrates the effect of $\beta$ (from 0.0 to 4.0) on MAGIC Search. Recall from Eq. (\ref{eq:score}) that, when $\beta=0.0$, the visual control (i.e., magic score in Eq. (\ref{eq:score})) is disabled, therefore MAGIC Search degenerates to the vanilla contrastive search~\cite{DBLP:journals/corr/abs-2202-06417}. From the results, we see that, by increasing $\beta$ from 0.0 to 0.5, a significant performance improvement in CIDEr is obtained. Such performance gain clearly demonstrates that the magic score in MAGIC Search is the key factor that enables the language model to generate text grounded on the given image. When $\beta$ is within the range of $[0.5,2.0]$, the performances of MAGIC Search are relatively the same, indicating the robustness of our approach in terms of the choice of $\beta$. On the other hand, we also see that a large $\beta$ (i.e., $\beta\geq2.5$) causes the performances to drop, suggesting that the importance of different terms (i.e., model confidence, degeneration penalty, and magic score in Eq. (\ref{eq:score})) in MAGIC Search should be properly balanced. In our experiments, the optimal setup for $\beta$ is 2.0.

\section{Detailed Results of Zero-Shot Image Captioning}
\label{appendix:zero_shot_image_captioning}

\begin{table*}[h]
    \small
	\centering  % 表居中
	\renewcommand{\arraystretch}{1.2}
	\setlength{\tabcolsep}{6pt}
	\scalebox{0.78}{
	\begin{tabular}{cccccccccccccc}
		\hlinewd{0.75pt}
		\multicolumn{14}{c}{\textbf{In-domain Result}}\\
		\hlinewd{0.75pt}
		\multirow{2}{*}{\textbf{Method}}&\multirow{2}{*}{run}&\multicolumn{6}{c}{MS-COCO}&\multicolumn{6}{c}{Flickr30k}\\
		\cmidrule(lr){3-8}
		\cmidrule(lr){9-14}
		&&B@1&B@4&METEOR&R-L&CIDEr&SPICE&B@1&B@4&METEOR&R-L&CIDEr&SPICE\\
		\hline
		\multirow{5}{*}{Top-$k$}&run-1&33.8&2.4&8.4&25.7&3.9&1.8&34.1&3.1&9.0&24.4&3.3&2.8\\
		&run-2&33.7&2.5&8.4&25.6&4.0&1.7&34.4&2.8&9.1&24.8&3.3&2.7\\
		&run-3&33.4&2.2&8.2&25.6&3.6&1.6&33.4&2.9&8.9&23.9&3.2&2.7\\
		\cmidrule(lr){2-14}
		&average&33.6&2.4&8.3&25.6&3.8&1.7&34.0&2.9&9.0&24.4&3.3&2.7\\
		&std&0.2&0.1&0.1&0.0&0.2&0.1&0.4&0.1&0.1&0.4&0.0&0.0\\
		\hline
		\multirow{5}{*}{Nucleus}&run-1&32.6&2.3&7.8&24.8&3.2&1.5&32.5&2.5&8.4&23.5&2.7&2.4\\
		&run-2&32.5&2.3&7.8&24.8&3.1&1.4&32.6&2.4&8.1&23.2&2.6&2.5\\
		&run-3&32.7&2.2&7.9&24.9&3.0&1.3&32.6&2.3&7.9&23.4&2.3&2.3\\
		\cmidrule(lr){2-14}
		&average&32.6&2.3&7.8&24.8&3.1&1.4&32.6&2.4&8.1&23.4&2.5&2.4\\
		&std&0.1&0.0&0.0&0.0&0.1&0.1&0.0&0.1&0.2&0.1&0.2&0.1\\
		\hlinewd{0.75pt}
		\multicolumn{14}{c}{\textbf{Cross-domain Result}}\\
        \hlinewd{0.75pt}
		\multirow{2}{*}{\textbf{Method}}&\multirow{2}{*}{run}&\multicolumn{6}{c}{MS-COCO $\Longrightarrow$ Flickr30k}&\multicolumn{6}{c}{Flickr30k $\Longrightarrow$ MS-COCO}\\
		\cmidrule(lr){3-8}
		\cmidrule(lr){9-14}
		&&B@1&B@4&METEOR&R-L&CIDEr&SPICE&B@1&B@4&METEOR&R-L&CIDEr&SPICE\\
		\hline
		\multirow{5}{*}{Top-$k$}&run-1&34.6&2.1&7.3&24.0&2.2&1.7&29.9&1.7&8.4&23.6&2.4&1.7\\
		&run-2&35.2&2.5&7.5&24.2&2.3&1.7&30.0&1.8&8.5&23.6&2.6&1.7\\
		&run-3&35.0&2.6&7.6&24.5&2.5&1.8&30.0&1.8&8.5&23.6&2.6&1.7\\
		\cmidrule(lr){2-14}
		&average&34.9&2.4&7.5&24.2&2.3&1.7&30.0&1.8&8.5&23.6&2.5&1.7\\
		&std&0.2&0.2&0.1&0.2&0.1&0.0&0.0&0.0&0.0&0.0&0.1&0.0\\
		\hline
		\multirow{5}{*}{Nucleus}&run-1&33.3&1.8&7.0&23.3&1.6&1.4&29.0&1.6&8.0&22.9&2.2&1.6\\
		&run-2&33.4&1.8&7.1&23.3&2.0&1.4&29.1&1.6&7.9&22.8&2.1&1.6\\
		&run-3&33.5&1.5&6.9&23.4&1.9&1.2&29.1&1.6&8.0&22.9&2.0&1.5\\
		\cmidrule(lr){2-14}
		&average&33.4&1.7&7.0&23.3&1.8&1.3&29.1&1.6&8.0&22.9&2.1&1.6\\
		&std&0.1&0.1&0.1&0.0&0.2&0.1&0.0&0.0&0.0&0.0&0.1&0.0\\
		\hlinewd{0.75pt}
	\end{tabular}}
    \caption{Complete numerical results of stochastic methods on zero-shot image captioning. The average and std rows show the mean and standard deviation of results from three different runs.}
    	\vspace{-1.5mm}
	\label{tb:complete_domain_stochastic_method_result}
\end{table*}

In Table \ref{tb:complete_domain_stochastic_method_result}, we show the detailed numerical results of stochastic baselines (i.e., Top-$k$ and Nucleus) on the task of zero-shot image captioning. The upper part of Table \ref{tb:complete_domain_stochastic_method_result} presents the results for in-domian experiments (Section \cref{sec:image_caption_result}) and the lower part of Table \ref{tb:complete_domain_stochastic_method_result} presents the results for cross-domain experiments (Section \cref{sec:cross_domain_image_captioning}). For each method, we report the results of three different runs with different random seeds along with the mean and standard deviation of different runs.

\clearpage

\section{More Visual Examples of Zero-Shot Image Captioning}
\label{appendix:image_captioning}

\begin{figure*}[hb] 
  \centering    
  \setlength{\abovecaptionskip}{3pt}
  \includegraphics[width=1.0\textwidth]{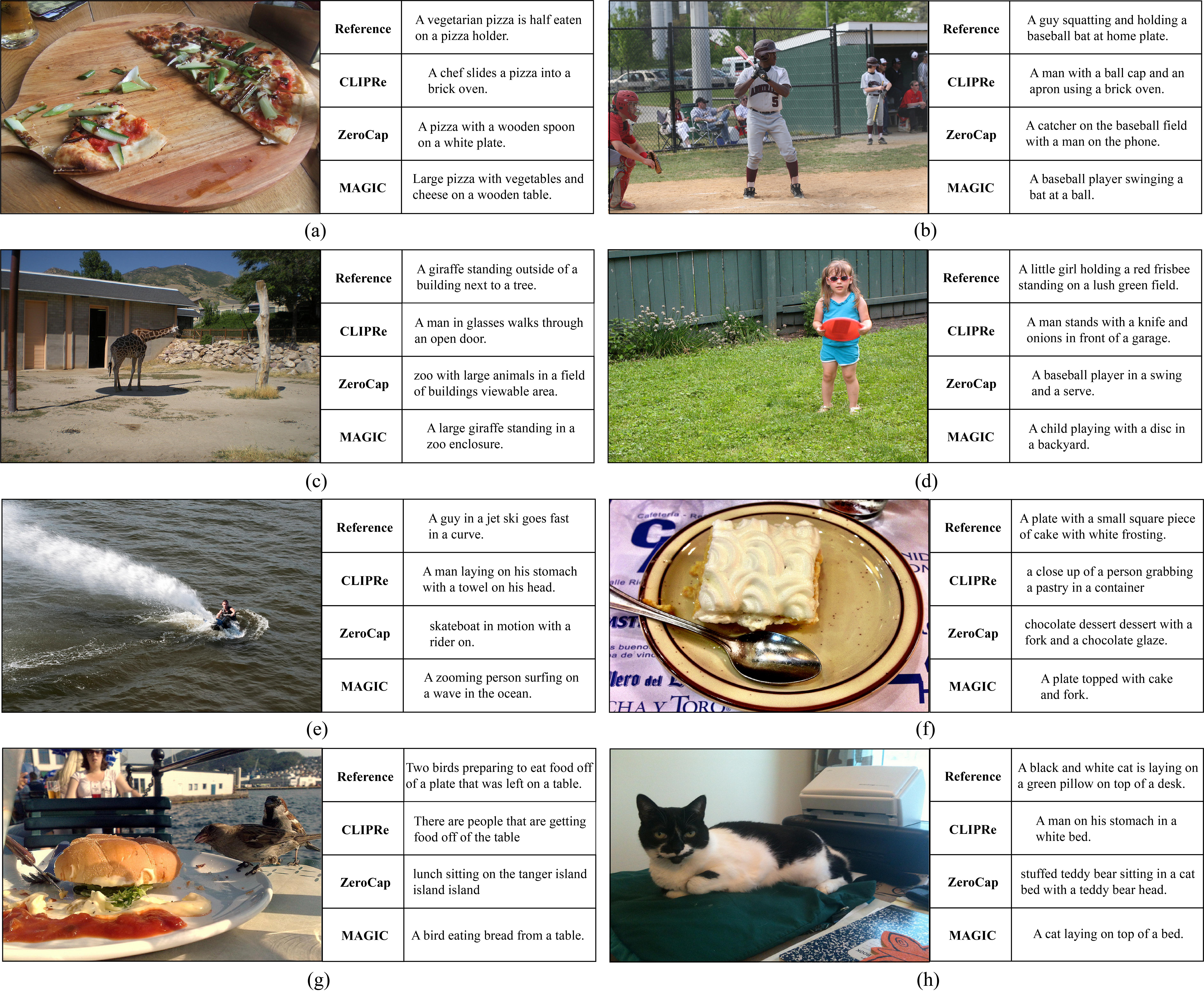}
  \caption{More examples of zero-shot image captioning. (Best viewed by zooming in.)}
  \label{fig:image_caption_case_study_appendix}
  \vspace{-1.5mm}
\end{figure*}

Figure \ref{fig:image_caption_case_study_appendix} presents more visual comparisons between our approach against other two strong zero-shot baselines (i.e., CLIPRe and ZeroCap) along with the reference caption. 

\section{Detailed Results of Story Generation}
\label{appendix:visually_grounded_story_generation}

\begin{table*}[tb]
    \small
	\centering  % 表居中
	\renewcommand{\arraystretch}{1.2}
	\setlength{\tabcolsep}{6pt}
	\scalebox{0.9}{
	\begin{tabular}{ccccccccc}
		\hlinewd{0.75pt}
		\textbf{Method}&run&rep-2$\downarrow$&rep-3$\downarrow$&rep-4$\downarrow$&diversity$\uparrow$&coherence$\uparrow$&MAUVE$\uparrow$&CLIPScore$\uparrow$\\
		\hline
		\multirow{5}{*}{Top-$k$}&run-1&3.42&0.75&0.22&0.95&0.460&0.85&0.21\\
		&run-2&3.27&0.77&0.25&0.95&0.455&0.87&0.21\\
		&run-3&3.46&0.75&0.21&0.95&0.458&0.86&0.21\\
		\cmidrule(lr){2-9}
		&average&3.38&0.76&0.23&0.95&0.458&0.86&0.21\\
		&std&0.08&0.01&0.02&0.00&0.002&0.01&0.00\\
		\hline
		\multirow{5}{*}{Nucleus}&run-1&2.88&0.57&0.16&0.96&0.448&0.88&0.21\\
		&run-2&2.88&0.61&0.18&0.96&0.454&0.87&0.22\\
		&run-3&2.99&0.62&0.20&0.96&0.454&0.88&0.21\\
		\cmidrule(lr){2-9}
		&average&2.92&0.60&0.18&0.96&0.452&0.88&0.21\\
		&std&0.05&0.02&0.02&0.00&0.003&0.00&0.00\\
		\hline
		\multirow{5}{*}{Typical}&run-1&2.44&0.42&0.11&0.97&0.454&0.83&0.18\\
		&run-2&2.56&0.48&0.12&0.97&0.448&0.84&0.19\\
		&run-3&2.55&0.48&0.14&0.97&0.448&0.85&0.19\\
		\cmidrule(lr){2-9}
		&average&2.52&0.46&0.12&0.97&0.450&0.84&0.19\\
		&std&0.05&0.03&0.01&0.00&0.002&0.01&0.00\\
		\hlinewd{0.75pt}
	\end{tabular}}
    \caption{Complete numerical results of stochastic methods on story generation. The average and std rows show the mean and standard deviation of results from three different runs.}
    	\vspace{-1.5mm}
	\label{tb:story_generation_complete_result}
\end{table*}

Table \ref{tb:story_generation_complete_result}, we present the detailed numerical results of stochastic baselines (i.e., Top-$k$, Nucleus, and Typical) on the task of story generation. For each method, we report the results of three different runs with different random seeds along with the mean and standard deviation of different runs.

\section{Human Evaluation Guidelines}
\label{appendix:evaluation_guideline}

Given the story title and the image, please evaluate the system's result with respect to the following features: (1) Coherence; (2) Fluency; (3) Informativeness; and (4) Story-Image Relevance. In the following, we provide some guidelines regarding how to judge the quality of the system's result in terms of different features.

\subsection{Coherence}
This metric measures whether the system's result is semantically and factually consistent with the story title. The definitions of different scores are:
\begin{itemize}
    \item \textbf{[5]}: The system's result is perfectly in line with the semantic meaning defined by the story title. And all its content is factually supported by or can be logically inferred from the title.
    \item \textbf{[4]}: The system's result is very related to the story title but with some minor errors that does not affect its overall relevance with respect to the story title.
    \item \textbf{[3]}: The system's result is, to some extent, relevant to the story title with some errors that display minor semantic inconsistency or contradiction. 
    \item \textbf{[2]}: At the first glance, the system's result seems to be related to the story title. But with careful inspection, the semantic inconsistency can be easily spotted.
    \item \textbf{[1]}: The system's result is obviously off-the-topic or it is semantically contradicted to the content contained in the story title. 
\end{itemize}

\subsection{Fluency}
This metric measures the fluency of the system's result. The definitions of different scores are:
\begin{itemize}
    \item \textbf{[5]}: The system's result is human-like, grammatically correct, and very easy to understand.
    \item \textbf{[4]}: Choose this score when you are hesitant between the score 3 and score 5.
    \item \textbf{[3]}: The system's result contains minor errors but they do not affect your understanding.
    \item \textbf{[2]}: Choose this score when you are hesitant between the score 1 and score 3.
    \item \textbf{[1]}: The system's result does not make sense and it is unreadable.
\end{itemize}

\subsection{Informativeness}
This metric measures the diversity, informativeness, and interestingness of the system's result. The definitions of different scores are:
\begin{itemize}
    \item \textbf{[5]}: The system's result is very informative and contains novel content. In addition, it displays a high level of diversity and it is enjoyable to read.
    \item \textbf{[4]}: Choose this score when you are hesitant between the score 3 and score 5.
    \item \textbf{[3]}: The system's result contains some new information and it displays a certain level of diversity.
    \item \textbf{[2]}: Choose this score when you are hesitant between the score 1 and score 3.
    \item \textbf{[1]}: The system's result is dull, repetitive, and does not have new information.
\end{itemize}

\subsection{Story-Image Relevance}
This metric measures the relevance between the system's result with respect to the given image. The definitions of different scores are:
\begin{itemize}
    \item \textbf{[5]}: The system's result is very related to the given image and contains a diverse set of concepts that are visually displayed in the image.
    \item \textbf{[4]}: Choose this score when you are hesitant between the score 3 and score 5.
    \item \textbf{[3]}: The system's result is partially related to the image and some of its content can be found in the image. 
    \item \textbf{[2]}: Choose this score when you are hesitant between the score 1 and score 3.
    \item \textbf{[1]}: The system's result is completely irrelevant to the given image. 
\end{itemize}

%\textbf{Participant Compensation.} To conduct the human evaluation, we hire 5 annotators in total and each annotator is paid by \$400. 

\section{More Examples of Story Generation}
\label{appendix:story_generation}
\begin{figure*}[hb] 
  \centering    
  \setlength{\abovecaptionskip}{3pt}
  \includegraphics[width=1.0\textwidth]{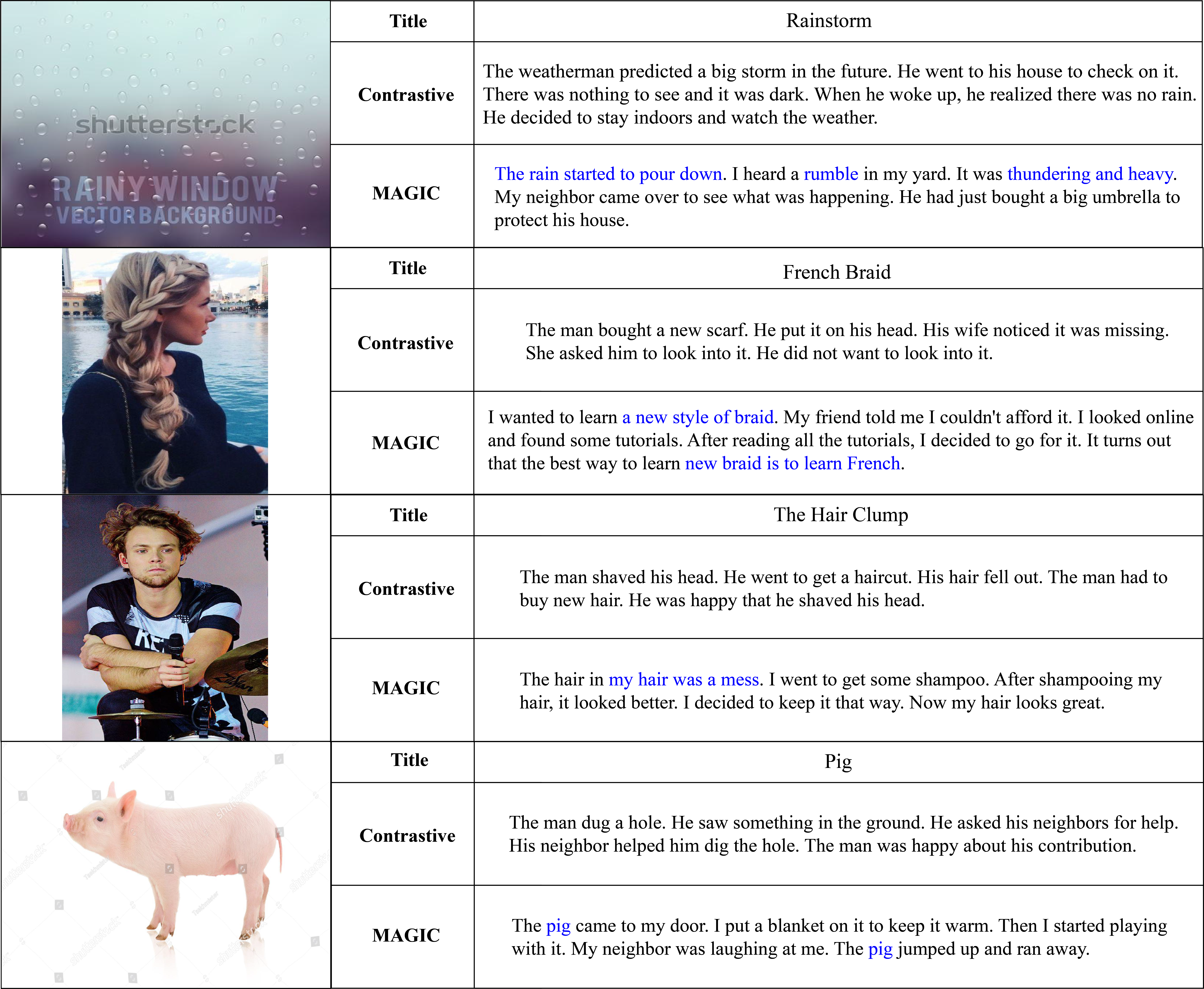}
  \caption{More examples of story generation. MAGIC can generate text (highlighted in \textcolor{blue}{blue}) that is related to the visual concepts displayed in the image. (Best viewed in color and by zooming in.)}
  \label{fig:story_generation_case_study_appendix}
  \vspace{-1.5mm}
\end{figure*}

Figure \ref{fig:story_generation_case_study_appendix} presents more examples generated by contrastive search along with the examples generated by MAGIC Search grounded on the retrieved image. The image retrieved by the story title is shown on the left-hand side of Table \ref{fig:story_generation_case_study_appendix}.

\end{document}